%% file: 0_main.tex
\documentclass[10pt, a4paper]{article}

\usepackage[final]{lrec2026} % this is the new style
\usepackage{amsmath}
\usepackage{booktabs}
\usepackage{array}
\usepackage{multirow}
\usepackage{colortbl}
\usepackage{subcaption}
\usepackage{comment}

\usepackage{tcolorbox}
\tcbuselibrary{listings,breakable}
\newtcblisting{promptbox}[1][]{
  colback=gray!5,
  colframe=black!60,
  boxrule=0.6pt,
  arc=4pt,
  fonttitle=\bfseries,
  title=#1,
  listing only,
  breakable,
  listing options={
    basicstyle=\ttfamily\footnotesize,
    breaklines=true,
    breakatwhitespace=true,
    columns=fullflexible,
    keepspaces=true,
        literate=
      {é}{{\'e}}{1}
      {è}{{\`e}}{1}
      {ê}{{\^e}}{1}
      {ë}{{\"e}}{1}
      {à}{{\`a}}{1}
      {â}{{\^a}}{1}
      {ù}{{\`u}}{1}
      {û}{{\^u}}{1}
      {ü}{{\"u}}{1}
      {ô}{{\^o}}{1}
      {î}{{\^i}}{1}
      {ï}{{\"i}}{1}
      {ç}{{\c{c}}}{1}
      {É}{{\'E}}{1}
      {È}{{\`E}}{1}
      {Ê}{{\^E}}{1}
      {À}{{\`A}}{1}
      {Ç}{{\c{C}}}{1}
      {œ}{{\oe}}{1}
      {Œ}{{\OE}}{1}
      {«}{{\guillemotleft}}{1}
      {»}{{\guillemotright}}{1}
      {—}{{---}}{1}
      {…}{{...}}{1},
  }
}

\newcommand{\tval}{0.7}
\newcommand{\nMHqueries}{885} % AUCUNE question removed
\newcommand{\nSHqueries}{897}

\newcommand{\nTotalQuestions}{1782}

\newcommand{\vspacea}{-0.3cm}
\newcommand{\vspacesmall}{0cm}
\newcommand{\vspaceb}{-0.5cm}
\newcommand{\azer}{0.85}

\title{HistoriQA-ThirdRepublic: Multi-Hop Question Answering Corpus for Historical Research, Parliamentary Debates from the French Third Republic (1870-1940)}

\name{Aurélien Pellet, Julien Perez, Marie Puren} 

\address{Affiliation1, Affiliation2, Affiliation3 \\
         Address1, Address2, Address3 \\
         author1@xxx.yy, author2@zzz.edu, author3@hhh.com\\
         \{author1, author5, author9\}@abc.org\\}

\name{Aurélien Pellet$^{1,2}$ \quad Julien Perez$^{3}$ \quad Marie Puren$^{1,4}$}

\address{
$^{1}$ LRE, EPITA, $^{2}$ EPITECH, $^{3}$ Bpifrance, $^{4}$ CJM\\
\{aurelien.pellet, marie.puren\}@epita.fr, julien.perez@bpifrance.fr\\}

\abstract{
We present HistoriQA-ThirdRepublic: a French-language dataset of multi-hop historical questions derived from parliamentary debates and newspapers of the French Third Republic. Designed in collaboration with a historian, the corpus captures complex reasoning patterns typical of historical inquiry, including cross-source synthesis, temporal reasoning, and the integration of sparse evidence. The dataset is made of \nTotalQuestions ~\ questions and emphasizes multi-hop connections across heterogeneous historical documents, providing a resource for evaluating retrieval-augmented and large language model systems in domain-specific contexts. We describe the methodology for constructing the corpus, including the selection and alignment of sources, question validation, and metadata integration. While the dataset focuses on French historical documents, our methodology can be readily adapted to other languages and national corpora. Finally, we demonstrate how the corpus can support realistic evaluation scenarios for multi-hop question answering, bridging the gap between NLP benchmarks and the needs of historical scholarship. 
\\ \newline 
}

\begin{document}

\maketitleabstract

\input{content/1_intro}
\input{content/2_related_work}
\input{content/3_corpus_description}

\input{content/4_methodology}
\input{content/5_experiments}
\input{content/6_conclusion}

\section{Bibliographical References}\label{sec:reference}
\bibliographystyle{lrec2026-natbib}
\bibliography{lrec2026-example}

\section{Language Resource References}\label{sec:lr-reference}
\bibliographystylelanguageresource{lrec2026-natbib}
\bibliographylanguageresource{lrec2026-example}

\input{content/7_appendix_error_analysis}

\end{document}

%% file: content/1_intro.tex
\begin{figure*}[t]
  \centering
  \includegraphics[width=\textwidth]{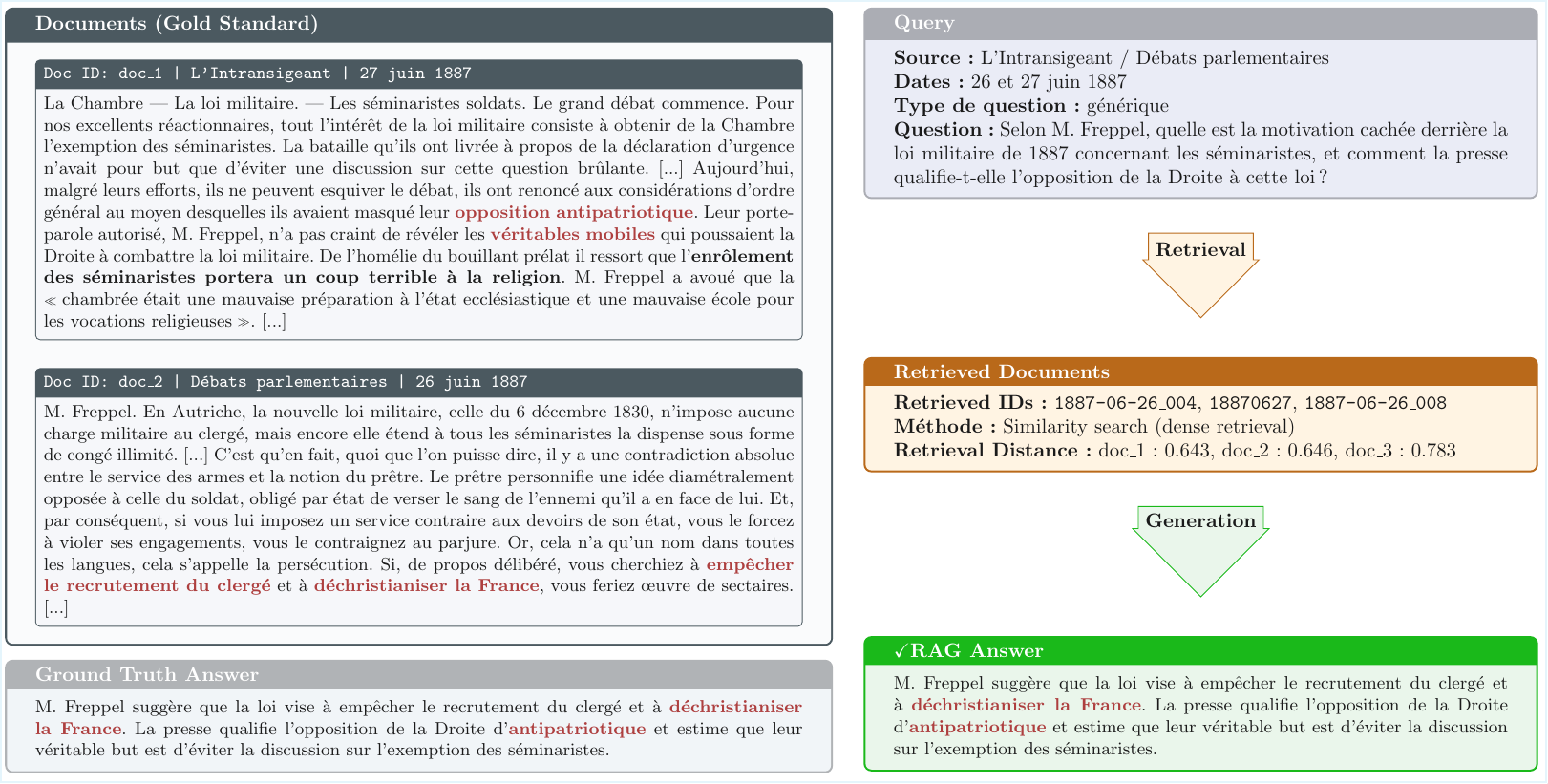}
  \caption{A sample of multi-hop question generated with Retrieval-Augmented Generation (RAG) pipeline. Given a query about the 1887 military law, the system retrieves relevant documents from heterogeneous sources and generates a multi-hop answer, which can be compared against the ground truth.}
  \label{fig:teaser}
\end{figure*}

\section{Introduction}

Large language models (LLMs) combined with retrieval-augmented generation (RAG) systems \cite{Lewis_Perez_Piktus_Petroni_Karpukhin_Goyal_Küttler_Lewis_Yih_Rocktäschel_et_al._2020} have recently achieved strong results on established benchmarks. A major challenge for LLMs lies in their training data \cite{guo2024biaslargelanguagemodels}. These models are rarely exposed to highly specific and hard-to-access documents, such as historical sources which, when digitized, are often stored in digital libraries. This limitation significantly impacts their ability to process and reason about specialized historical corpora, where domain expertise and contextual understanding are paramount. By grounding generation in retrieved documents, RAG reduces—though does not eliminate—hallucination and, in some cases, enables good performance without task-specific fine-tuning. As RAG architectures diversify \cite{Chang_Jiang_Rakesh_Pan_Yeh_Wang_Hu_Xu_Zheng_Das_et_al._2025,Li_He_Liu_Zhang_Yu_Ye_Zhu_Su_2025}, reported benchmark scores have continued to rise \cite{llama3,gpt4}, but these gains mask important limitations when models are applied to complex, domain-specific tasks.

First, hallucination and retrieval errors remain a problem in long-context and multi-document scenarios: when answers require reasoning over lengthy documents or the synthesis of information from several heterogeneous sources, errors compound, and factual consistency degrades. Even for RAG, it has been shown how failure can increase in long context \cite{Leng_Portes_Havens_Zaharia_Carbin_2024}.

Second, standard benchmarks remain poorly aligned with domain-specific challenges. They usually focus on multiple-choice and short-answer questions, short passages, or single-document question answering \cite{mmlu,Zhuang_Zhang_Cheng_Yang_Liu_Huang_Lin_Rajmohan_Zhang_Zhang_2024}. Recent initiatives have attempted to address this limitation by proposing more targeted benchmarks, such as Humanity’s Last Exam \cite{phan2025humanitysexam}. However, many of these datasets are still relatively general in scope and focus primarily on factual questions.

These shortcomings become particularly evident in the case of historical corpora, where existing benchmarks fail to capture the kinds of reasoning that domain experts require such as  temporal reasoning, cross-source linking, and the handling of highly specific, hard-to-access documents.To illustrate this, we focus on a case study of the French Third Republic (1870–1940), drawing on agenda-setting theory \cite{McCombs_Shaw_1972} to situate parliamentary debates within their media ecosystem. This period, marked by an autonomous Parliament \cite{Morel_2024} and the rise of the “grande presse d’information\footnote{A term coined to describe French major daily newspapers born in the second half of the 19th century.}” \cite{kalifa_trerenty_2011}, provides an ideal context for studying how complex, cross-source questions emerge. Fully digitized corpora of parliamentary debates\footnote{Available online for the years 1880–1940 for both the \href{https://gallica.bnf.fr/ark:/12148/cb328020951/date&rk=150215;2}{Chamber
 of Deputies} and the \href{https://gallica.bnf.fr/ark:/12148/cb34363182v/date&rk=128756;0}{Senate}.}
 exemplify these challenges: historians often need to connect evidence across heterogeneous documents, trace arguments over time, and aggregate sparse mentions dispersed across records. To better capture how issues circulated and gained prominence, we complement this corpus with digitized newspapers from the same period. By linking parliamentary proceedings with contemporary press coverage, we aim to reconstruct the dynamic interplay between institutional discourse and public framing, and to support the generation of questions that reflect the full complexity of historical agenda-setting processes. The issue lies not only in the thematic scope of the questions, but also in the types of reasoning that are relevant to historians. Those requirements affect every component of a RAG pipeline: indexing and chunking methodologies for long texts, retrieval architectures and similarity metrics, the exploitation of metadata and temporal signals, and answer-generation approaches that must support complex argumentative or interpretive responses rather than only fact verification or multiple-choice answers.

In this paper, we present a method for generating, analyzing, and validating high-quality questions that reflect historians’ research problems. Drawing on historical parliamentary debates (sources that are thematically rich, extensive, and span several decades) and cross-referencing them with contemporaneous press archives, we demonstrate how to incorporate temporal dynamics and source sparsity into question generation, and how clustering and metadata can help identify promising multi-hop candidates. We further analyze the structure of both the corpora and the questions to identify natural seeds for multi-hop question generation. We evaluate RAG systems at both the retrieval and generation stages on a carefully selected subcorpus of French Third Republic parliamentary proceedings.

Our work makes four main contributions. First, we propose HistoriQA-ThirdRepublic\footnote{Dataset available at \url{https://github.com/atomegoyan/historiqa-thirdrepublic}} a dataset historically aligned, requiring reasoning across diverse sources and formats, and designed to challenge RAG systems. It comprises \nSHqueries ~\ single-hop and \nMHqueries ~\ multi-hop questions, all automatically generated through controlled prompting and validated by a historian for factual and contextual coherence. Second, we introduce a method for generating domain-specific multi-hop questions from heterogeneous corpora, combining embedding-based similarity, temporal constraints, and expert-guided prompt engineering. We further evaluate the resulting questions by analyzing their evolution in semantic space and conducting a preliminary classification of their types. Third, we introduce an evaluation protocol that leverages historian judgments to assess both question and answer quality. Finally, we conduct an automatic evaluation of RAG retrieval and generation performance and investigate the alignment between LLM-based judgments and human annotations.

Taken together, these contributions aim to bridge the gap between off-the-shelf RAG benchmarks and the specific needs of historical scholarship, enabling more meaningful evaluation and targeted development of retrieval and generation methods for complex, domain-specific corpora in the humanities and social sciences.

%% file: content/2_related_work.tex
\section{Related work}

%The computational study of debates from the Third French Republic has already yielded significant insights. Methods such as topic modeling using LDA \cite{bourgeois:hal-03526254} or dense semantic embeddings like BERT \cite{puren:hal-04128262} have offered a broader perspective on the range of issues addressed in the French parliament.

%Automatic question generation has already shown promising results in this context. Recent studies have assessed the ability of retrievers to reason over long-context documents, focusing in particular on the comparison of chunking strategies and the evaluation of different retriever architectures \cite{pellet:hal-04832663,pellet:hal-05193494}. Earlier work relied on fixed-size chunks \cite{Smith_Troynikov_2024}, determined by the chosen segmentation strategy, but did not account for the intrinsic structure. As a result, some chunks may naturally be longer than others, and enforcing uniform segmentation can be problematic.

\paragraph{Multi-Document Reasoning Datasets}
Multi-document reasoning have been developed to challenge information retrieval and question answering systems. Resources such as HotpotQA~\cite{hotpotqa} and 2WikiMultihopQA~\cite{ho-etal-2020-constructing} provide deeper insights into the question-generation process. In particular, HotpotQA introduces a detailed taxonomy of question categories, covering various question types and forms of multi-hop reasoning. Building on this line of work, datasets like MoreHopQA~\cite{morehopqamultihopreasoning} propose even more complex multi-hop questions that require additional layers of reasoning, including commonsense, arithmetic, and symbolic inference.

\paragraph{Automatic Question Generation for RAG Evaluation}
Research on automatic question generation has shown promising results for improving the evaluation of RAG systems in specific context, particularly by enabling the systematic assessment of a model’s reasoning capabilities over long and structured documents. By combining embeddings, metadata, keywords, and clustering techniques, it is possible to generate domain-aligned questions ranging from single-hop factual queries to multi-hop, temporally grounded questions that require cross-source synthesis. Such synthetic question sets can be used to stress different parts of the RAG pipeline and to create more realistic evaluation scenarios for domain specialists \cite{Lin_Chen_Song_Zhang_2024,Li_Zhang_Kong_2025,Hwang_Kim_Lee_2024}.

\paragraph{Retriever Performance in Long-Context and Historical Documents}
Recent studies have examined the ability of retrievers to reason over long-context documents in a historical context with particular attention to the comparison of chunking strategies and the evaluation of different retriever architectures~\cite{pellet:hal-05193494}. Earlier work often relied on fixed-size chunks~\cite{Smith_Troynikov_2024}, determined by the chosen segmentation strategy, without accounting for the documents’ intrinsic structure. As a result, some chunks may naturally be longer than others, and enforcing uniform segmentation can be problematic.

%When dealing with parliamentary debates, a natural chunking strategy (such as splitting by speech turns) can result in chunks of highly variable length. A retriever returning $k$ documents for a given query may thus encounter substantial variability in input size: for one query, the retrieved passages may total 10,000 tokens, while for another, the same $k$ results may contain only 2,000 tokens. In addition, documents from newspaper corpora tend to be shorter on average, as shown in Table~\ref{tab:corpus-stats}. This discrepancy reduces the comparability of results across different $k$ values. To address this limitation, a refinement has been introduced: computing standard metrics such as recall@$k$—the proportion of times the gold document is retrieved among $k$ candidates—at the token level, resulting in a recall@Tokens metric calculated for various token thresholds.

%However, existing experiments have primarily focused on chunking strategies, leaving the issue of query generation largely unexplored. Only single-hop questions were generated, and evaluation efforts targeted exclusively the retriever component of RAG systems. Furthermore, question generation relied solely on debates from the French Third Republic, thereby restricting the LLM’s reasoning capabilities to a single type of source.

%\subsection{Historical Context}
%TO DO

%% file: content/3_corpus_description.tex
\section{Corpus Description}

The parliamentary debates of the French Third Republic are available as plain text files resulting from OCR, provided through the digital library of the Bibliothèque nationale de France. For this study, we focus on the 1887 debates in the Chamber of deputies, which provides continuous coverage of parliamentary sessions while remaining small enough for detailed analysis\footnote{For the Chamber of Deputies only, there are approximately 11,000 automatically transcribed pages for the 1885–1889 parliamentary term.}. In 1887, the publicity of parliamentary debates was also firmly established, allowing the press to report extensively on parliamentary proceedings. That same period was marked by international tensions (most notably with Germany), political scandals reaching the highest levels of the state, and a surge of anti-parliamentarism crystallized in the ``Boulangist crisis'', which posed a major challenge to the republican regime\footnote{General Boulanger, whose popularity continued to grow during his two terms as Minister of War in 1886 and 1887, subsequently assumed the leadership of a vigorous anti-parliamentary coalition.}. These events found in the press both a powerful amplifier and, in some cases, an active ally, as certain newspapers openly supported Boulanger’s followers. These crisis illustrate how parliamentary debates and press coverage became deeply intertwined, shaping public perception and fueling political polarization. This context provides a particularly rich case for studying the interplay between Parliament and media, as it captures both institutional discourse and its mediation in the public sphere.

We complement the corpus of parliamentary proceedings with two contemporary newspapers, \textit{Le Gaulois} and \textit{L’Intransigeant}, using plain-text transcripts released by the Bibliothèque nationale de France\footnote{Datasets are available via the \href{https://api.bnf.fr/index.php/fr/texte-des-documents-de-presse-du-projet-europeana-newspapers-xixe-xxe-siecles}{API and dataset portal} of the Bibliothèque nationale de France.}. These daily newspapers represent two opposing positions on the political spectrum: on the one hand, monarchist and conservative right-wing views; on the other, socialism, although \textit{L’Intransigeant} would gradually shift toward populism. This multimodal perspective is designed to test whether RAG can connect parliamentary discourse with press coverage, thus requiring reasoning across heterogeneous sources.

While this temporal focus inevitably narrows the scope of the corpus, it remains substantial enough to enable close collaboration with historians, who can provide qualitative feedback on generated questions and RAG outputs.

\begin{table*}[t]
\centering
\begin{tabular}{lcccc}
\toprule
 & \multicolumn{2}{c}{\textbf{Les Débats}} & \textbf{Le Gaulois} & \textbf{L’Intransigeant} \\
\cmidrule(lr){2-3}
 & {All documents} & {Filtered} &  &  \\
\midrule
Number of documents        & 3229 & 963 & 78  & 79 \\
Average tokens / document  & 2020 & 4433 & 922 & 376 \\
Median tokens / document   & 279 & 716 & 802 & 317 \\
Range (min–max)            & 1–63121 & 143–62998 & 51–4023 & 6–1335 \\
Standard deviation          & 6636 & 9697 & 675 & 257 \\
\bottomrule
\end{tabular}
\caption{Statistics of the corpus: \textit{Les Débats}, \textit{Le Gaulois}, and \textit{L’Intransigeant}. 
For \textit{Les Débats}, we report both the full dataset and a filtered subset excluding non-debate sections.}
\label{tab:corpus-stats}
\end{table*}

Table~\ref{tab:corpus-stats} reports descriptive statistics for the three corpora used in our study.
Each corpus consists of documents or chunks. For the parliamentary debates, segmentation relied on regular expressions identifying session boundaries and speaker turns, while enforcing a minimum chunk size of 10,000 characters. Newspaper articles were subsequently extracted via dedicated regex patterns applied to the relevant portions of the reports. The debates dataset (Les Débats) comprises more than 3,200 documents.

To enhance textual coherence, we filtered out non-debate sections (e.g., agendas, procedural notes, vote lists). The resulting 963 documents display wide variation in length, mirroring the thematic diversity and rhetorical dynamics of the assembly. This filtered subset thus offers a more faithful view of deliberative discourse while reducing noise for downstream analysis.

The two complementary corpora consist of press reports on parliamentary sessions published in \textit{Le Gaulois} and \textit{L’Intransigeant}\footnote{These reports, written by specialized journalists, summarized, sometimes in the form of verbatim transcripts, and commented on the debates held in the Chamber. For readers, they were often the primary source of information on parliamentary activity.}.

Although shorter than the parliamentary proceedings, these texts add interpretative and often polemical perspectives. Combined with the debates, they create a rich contrast between institutional discourse and mediated public interpretation—an essential feature for evaluating multi-hop reasoning.

%% file: content/4_methodology.tex
\section{Methodology}

%In this section, we describe our approach for generating and evaluating historical questions from our corpora. We distinguish between single-hop questions, which rely on a single text segment, and multi-hop questions, which require linking information across debates and newspapers. For each case, we detail the prompting strategies, embedding-based similarity measures, and visualization techniques used to assess the relation between generated questions and their source material.

In this section, we present our approach to generating and evaluating historical questions from the corpus. We describe the prompting strategies, embedding-based similarity measures, and visualization techniques used to assess the relation between the generated questions and their source material.

\subsection{Single-hop Question Generation}

We focus on generating questions answerable from single text chunks, restricting our scope to parliamentary debates from the filtered sub-corpus (963 documents). Through iterative refinement with our team historian, we developed a prompt designed to favour (i) substantive debate content over procedural detail, (ii) the articulation of political positions and key issues, and (iii) paraphrase over surface-level text reuse. The model returned structured JSON output, yielding an empty list when no suitable question could be produced.

\begin{figure}[htbp]
\centering
\includegraphics[width=\columnwidth]{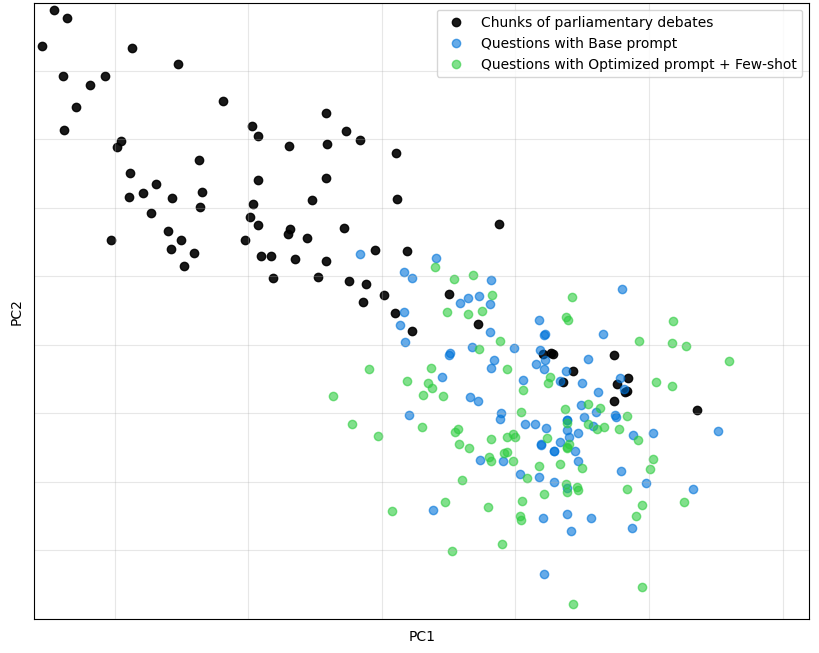}
\vspace{\vspaceb}
\caption{TF–IDF projection of generated questions and source texts. Questions from the optimized prompt lie farther from their source chunks than those from the base prompt, indicating less lexical overlap and greater complexity.}
\label{fig:pca_embeddings_questions_simplified}
\vspace{\vspacea}
\end{figure}

To assess semantic relationships, we computed TF–IDF representations and applied PCA dimensionality reduction (Figure~\ref{fig:pca_embeddings_questions_simplified}). On average, the cosine distance ($d = 1 - \text{similarity}$) between questions and their source chunks is 1.04 for the baseline prompt and 1.26 for the optimized prompt indicating higher semantic divergence after prompt refinement. This increase suggests that the optimized prompt encourages conceptual reformulation rather than lexical copying—a trend corroborated by qualitative inspection.

% Average cosine distances between questions and their source chunks were 1.04 (baseline) and 1.26 (optimized), indicating higher semantic divergence after prompt refinement. This increase suggests that the optimized prompt encourages conceptual reformulation rather than lexical copying—a trend corroborated by qualitative inspection.

\begin{comment}
\begin{table}[htbp]
\centering
\tiny
%\renewcommand{\arraystretch}{1.3}
\begin{tabular}{p{0.8cm}p{2.8cm}p{2.8cm}}
\toprule
\textbf{Date} & \textbf{Question (French)} & \textbf{English Translation} \\
\midrule
1887-05-23 & Quels sont les principaux points de désaccord entre le ministre des travaux publics, M. de Hérédia, et M. Jaurès concernant le projet de loi sur la sécurité des mineurs et la représentation des ouvriers dans les mines en 1887 ? & What are the main points of disagreement between the Minister of Public Works, M. de Hérédia, and M. Jaurès regarding the bill on miner safety and worker representation in mines in 1887? \\[1em]
\midrule
1887-11-24 & Quels projets de loi ont été adoptés en 1887 concernant les emprunts des villes de Sedan, Dijon et Niort, et quelle en est la raison ? & What bills were adopted in 1887 concerning the loans of the cities of Sedan, Dijon and Niort, and what was the reason? \\[1em]
\midrule
1887-10-29 & En 1887, quelle est la position de la Chambre des députés sur la demande d'urgence pour la loi concernant le traitement des instituteurs ? & In 1887, what is the position of the Chamber of Deputies on the urgent request for the law concerning teachers' salaries? \\[1em]
\bottomrule
\end{tabular}
\caption{Sample of single-hop questions generated from the corpus. French questions are paired with their English translations.}
\label{tab:singlehop_examples}
\end{table}
Table~\ref{tab:singlehop_examples} shows examples of single hop questions generated.
\end{comment}

\subsection{Multi-hop question generation}

\begin{figure}[htbp]
\centering
\includegraphics[width=\columnwidth]{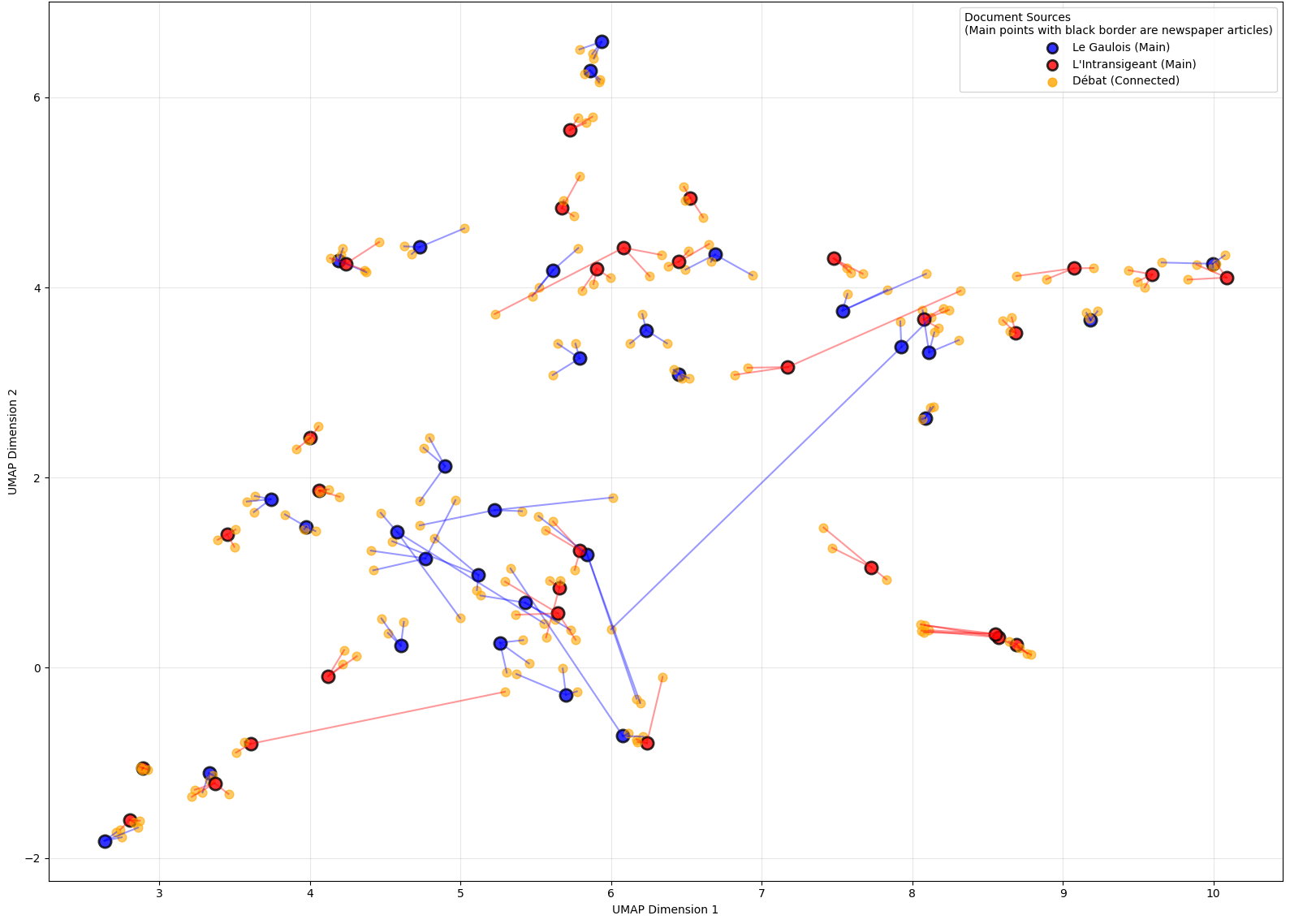}
\vspace{\vspaceb}
\caption{UMAP visualization of semantic relationships between press and debate articles. Each main point (black-bordered) represents an article from \textit{Le Gaulois} (blue) or \textit{L’Intransigeant} (red). Lines connect each article to its three most similar debate chunks (Cohere embeddings, cosine similarity). The 2D projection reveals cross-newspaper thematic links between press and parliamentary discourse.}
\label{fig:newspaper_cross_references}
\vspace{\vspacea}
\end{figure}

We next address the generation of multi-hop questions that span both parliamentary debates and newspaper articles. A natural starting point is to pair text chunks that are temporally proximate. While this approach substantially reduces the number of potential combinations, it does not ensure that the retrieved documents are semantically related, and thus suitable for multi-hop reasoning within a RAG framework.

Figure~\ref{fig:newspaper_cross_references} presents a two-dimensional projection of newspaper samples, each linked to its top-3 most similar debate chunks, highlighted in yellow. This visualization illustrates the potential for identifying candidates for multi-hop question generation: in many cases, clusters of semantically close text segments can be observed in the embedding space. The embedding model appears to capture meaningful relationships between press samples and parliamentary debates. This unsupervised visualization supports our assumption that appropriate candidates for multi-hop question generation can be detected automatically. To refine the selection, we applied a similarity threshold of $\tval$, ensuring that only highly relevant pairs were retained.
For the multi-hop question generation phase, we once again collaborated with the historian on our team to iteratively refine the prompting strategy. Our expert provided domain-specific examples that helped define the types of relationships and reasoning chains most valuable for historical interpretation.

We identified two complementary dimensions for guiding the generation process. The first concerns the \textit{type} of questions.
Inspired by the HotpotQA taxonomy of questions~\cite{hotpotqa}, early iterations produced a first set of multi-hop questions that involved, for instance, identifying an opinion expressed in the debates and examining how the press reacted to it; we refer to these as \textit{follow-up questions}. We then extended this to \textit{bridge-entity questions}, where a shared reference (e.g., a political figure or event) connects a debate passage with a newspaper article. Finally, we introduced \textit{comparative questions} designed to contrast differing viewpoints across multiple sources. The full prompts used for these experiments are detailed in Appendix~\ref{sec:appendix_prompts} and provided in the supplementary material available in the \href{https://github.com/atomegoyan/historiqa-thirdrepublic}{dataset} repository.

The second dimension relates to the \textit{source pairing strategy}. We focused on two types of document relationships:\\

\noindent\textbf{Newspaper-to-debate questions.} For each newspaper chunk, we retrieved its top-1 most similar debate passage based on cosine similarity between embeddings. This direction (press to debate) was chosen to limit the total number of generated questions while maximizing the diversity of reasoning types. A similarity threshold of $\tval$ was again applied to filter weakly related candidates.\\

\noindent\textbf{Newspaper-to-newspaper questions.} We computed all similarity pairs between the two selected newspapers. A threshold of $\tval$ was applied, along with a temporal constraint requiring that the paired articles be published no more than one week apart.

For the present study, we set aside multi-hop questions relying exclusively on parliamentary debates. Although such questions could provide insights into the evolution of political arguments over time, they carry the risk that multi-hop reasoning would occur across adjacent or even identical sessions, thereby limiting the diversity of perspectives. By combining debates with newspapers, we instead emphasize the interpretive gap between official parliamentary proceedings and their representation in the press. This approach introduces an additional challenge: the retrieval process must operate across distinct corpora, each characterized by different styles, formats, and lengths. 

\begin{figure*}[htbp]
\centering

\begin{subfigure}[t]{0.32\textwidth}
    \centering
    \includegraphics[width=\linewidth]{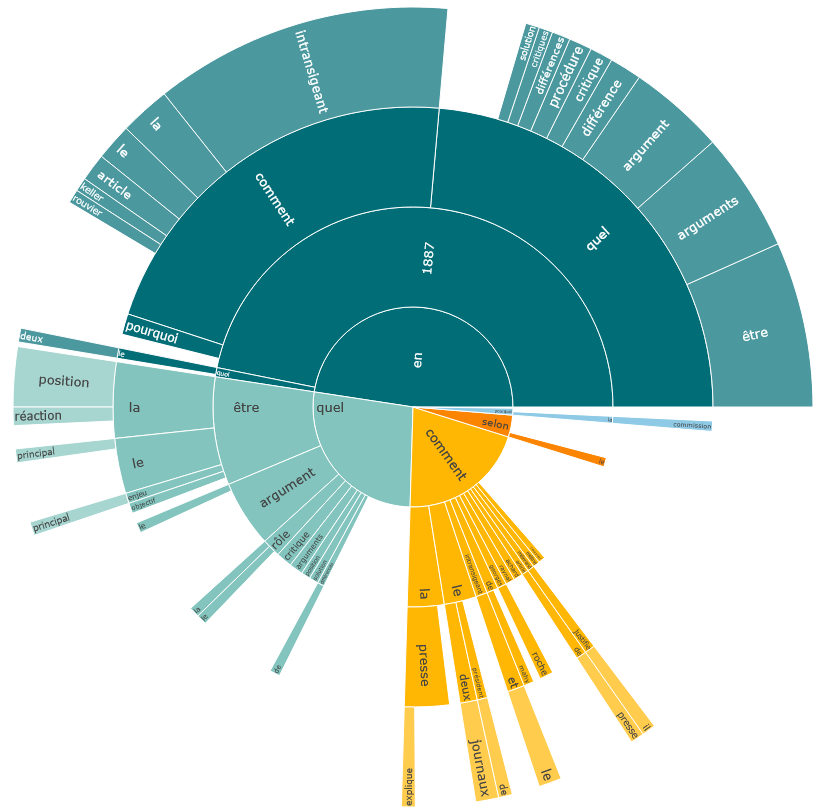}
    \caption{Follow-up questions.}
    \label{fig:first_figure}
\end{subfigure}
\hfill
\begin{subfigure}[t]{0.32\textwidth}
    \centering
    \includegraphics[width=\linewidth]{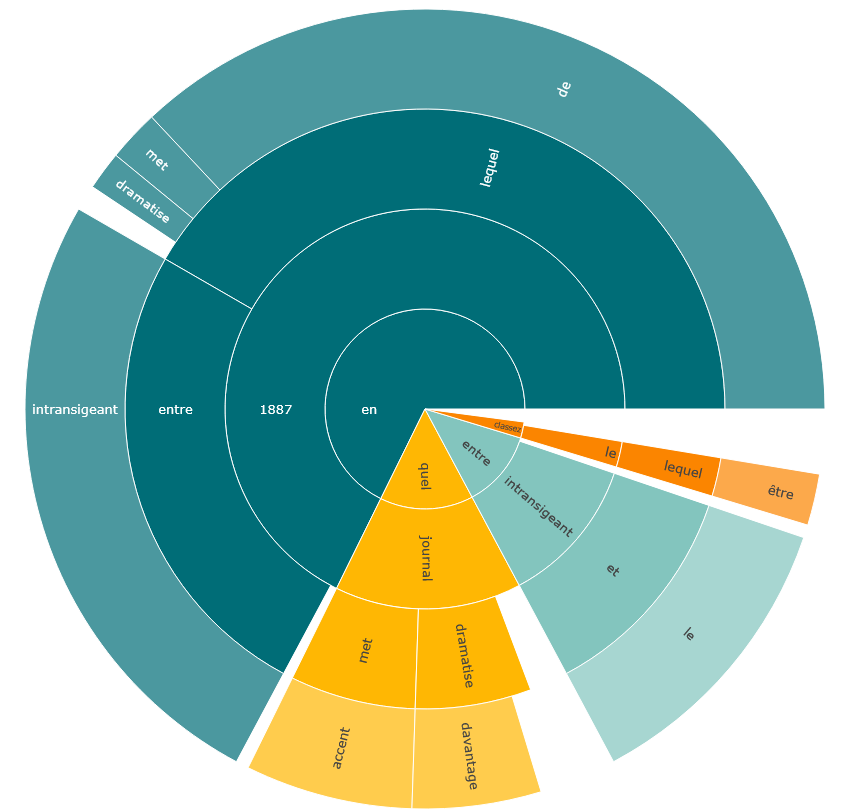}
    \caption{Comparative questions}
    \label{fig:second_figure}
\end{subfigure}
\hfill
\begin{subfigure}[t]{0.32\textwidth}
    \centering
    \includegraphics[width=\linewidth]{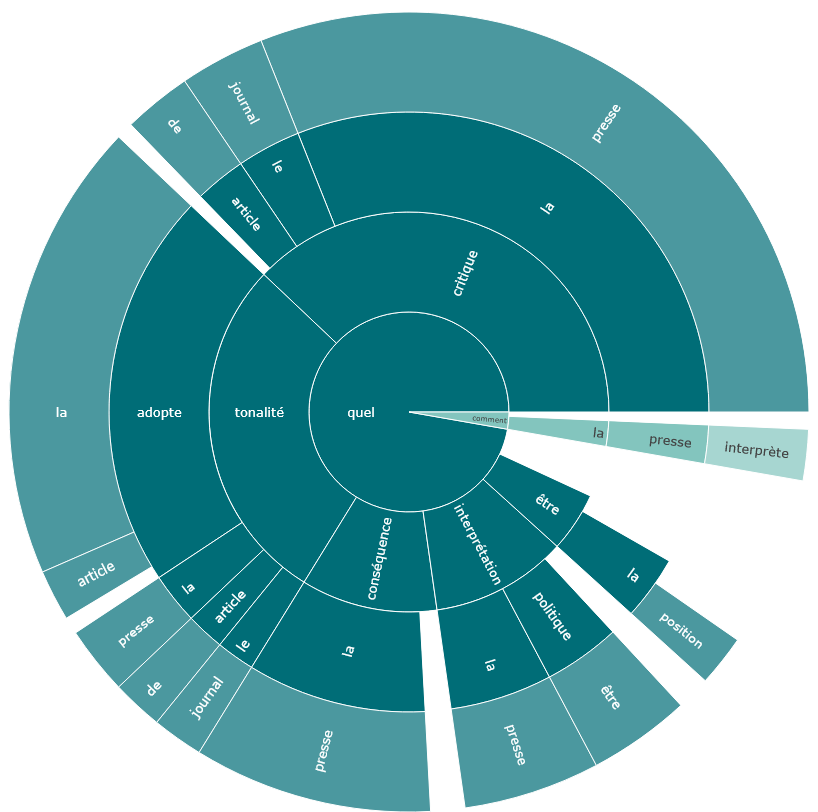}
    \caption{Bridge-entity questions}
    \label{fig:third_figure}
\end{subfigure}

\caption{Details of the generated questions, illustrating frequency patterns for each category. Each segment represents a possible word continuation in the sequence of question prefixes (lemmatized and in French). Empty colored blocks indicate rare suffixes not displayed individually. Subfigure~(a) corresponds to follow-up questions, (b) to comparative questions, and (c) to bridge-entity questions.}
\label{fig:comparison}
\end{figure*}

Table~\ref{tab:multihop_examples} presents samples of the generated questions. In total, we produced $\nMHqueries$ multi-hop questions. Figure~\ref{fig:comparison} presents a sunburst plot for the first four words of the generated questions.

\begin{table}[htbp]
\centering
\tiny
\begin{tabular}{p{0.4cm}p{2cm}p{2cm}p{2cm}}
\toprule
\textbf{Date} & \textbf{Question (French)} & \textbf{English Translation} & \textbf{Type / Sources} \\
\midrule
1887 & En 1887, quelle est la position de L'Intransigeant et de Le Gaulois concernant l'intégration des individus nés en France de parents étrangers dans le service militaire, et comment ces positions reflètent-elles leurs visions respectives de la nationalité et de l'armée ? & In 1887, what is the position of L'Intransigeant and Le Gaulois regarding the integration of individuals born in France to foreign parents into military service, and how do these positions reflect their respective views on nationality and the army? & Follow-up / Le Gaulois + L'Intransigeant \\[1em]
\midrule
1887 & En 1887, comment la presse critique-t-elle le processus de création de chaires d'enseignement supérieur, et en quoi cela contraste-t-il avec les arguments du ministre de l'Instruction publique lors du débat parlementaire ? & In 1887, how does the press criticize the process of creating university chairs, and how does this contrast with the arguments of the Minister of Public Instruction during the parliamentary debate? & Follow-up / Le Gaulois + Les Débats \\[1em]
\midrule
1887 & Quelle tonalité la presse adopte-t-elle à propos du député qui a défendu une augmentation des crédits pour l'agriculture face à la concurrence étrangère ? & What tone does the press adopt regarding the deputy who defended an increase in credits for agriculture in the face of foreign competition? & Bridge-entity / Les Débats + L'Intransigeant \\[1em]
\bottomrule
\end{tabular}
\caption{Sample of generated multi-hop questions from the corpus. French questions are paired with their English translations, and the last column indicates the question type and associated sources.}
\label{tab:multihop_examples}
\vspace{\vspacea}
\end{table}

%% file: content/5_experiments.tex
\section{Experiments}

%\subsection{Research Questions}

This section addresses three main questions: 
How do current models perform on a long-context historical dataset ? (RQ1)
To what extent do LLM-based evaluations of retrieval and generation align with human judgments ? (RQ2)
How are question types related to model performance and error patterns ? (RQ3)

\vspace{\vspacesmall}

\subsection{Experimental Settings}

Question generation experiments were conducted using the Cohere API. We employed the \texttt{Command-R} model for reasoning-oriented tasks and \texttt{Command-A} for both question generation and RAG-based answering. Dense retrieval was performed using embeddings computed with \texttt{cohere-embed-v4}, while retrieved passages were reranked with \texttt{cohere-rerank-multilingual-v3.0}. The vector database was implemented using ChromaDB. 

In addition to Cohere models, we evaluated several large language models for both generation and RAG-based answering, including \texttt{Llama~3.2~3B~Instruct}, \texttt{Llama~3.3~70B~Instruct}, \texttt{GPT-OSS~20B}, and \texttt{GPT-4o-mini}.

The final prompts and system configuration were refined through multiple iterative cycles in close collaboration with our team historian, who is a specialist of this period. This process enabled the selection of representative few-shot examples for question generation, LLM-as-a-judge evaluation, and RAG-based answering.

\begin{comment}
\subsection{Single-Hop vs Multi-Hop Question Classification}

We employed a logistic regression classifier to distinguish between single-hop and multi-hop questions, using Cohere embeddings as latent representations. The model achieved high performance on this binary classification task. 

\begin{table}[htbp]
\centering
\footnotesize
\setlength{\tabcolsep}{5pt}
\renewcommand{\arraystretch}{1.4}
\begin{tabular}{cc|cc}
\toprule
& & \multicolumn{2}{c}{\textbf{Predicted}} \\
& & \textbf{Single-Hop} & \textbf{Multi-Hop} \\
\midrule
\multirow{2}{*}{\rotatebox{90}{\textbf{Actual}}}
& \textbf{Single-Hop} & 494 & 1 \\
& \textbf{Multi-Hop}  & 5   & 265 \\
\bottomrule
\end{tabular}
\caption{
Confusion matrix for the \textbf{Single-Hop vs. Multi-Hop} classifier.  
Rows correspond to \textbf{actual (human) labels}, and columns to \textbf{model predictions}.
}
\label{tab:confusion_matrix_singlehop}
\end{table}

Table~\ref{tab:confusion_matrix_singlehop} shows that this simple classifier achieves near-perfect performance, with a test accuracy of 99.22% and minimal signs of overfitting. The confusion matrix reports only 6 misclassifications out of 765 test samples: 5 multi-hop questions incorrectly classified as single-hop and 1 single-hop question misclassified as multi-hop.
\end{comment}

\begin{table*}[ht]
\centering
\footnotesize
\begin{tabular}{lcc}
\toprule
\multicolumn{1}{c}{\textbf{Model}} & \multicolumn{2}{c}{\textbf{Accuracy}} \\
\cmidrule(lr){1-1} \cmidrule(lr){2-3}
\textbf{Model} & \textbf{Single-hop} & \textbf{Multi-hop} \\
\midrule
\texttt{Llama-3.3-3b-Instruct} & 24.10 & 3.2 \\
\texttt{Llama-3.3-70b-Instruct} & 53.53 & 17.92 \\
\midrule
\texttt{gpt-oss-20b} & \textbf{58.85} & \textbf{18.86} \\
\texttt{gpt-4o-mini} & 52.72 & 13.77 \\
\bottomrule
\end{tabular}
\caption{Evaluation on Single-hop and Multi-hop datasets, using accuracy as judged by an LLM evaluator. Larger models show higher performance across both datasets, especially on the Multi-hop benchmark.}
\label{tab:llm_eval_clean}
\end{table*}

\subsection{Retrieval Evaluation}

We evaluate the retrieval stage of the RAG pipeline by measuring how often the gold source chunks used to generate each question are successfully returned by the retriever. 

% Because questions were generated from known source passages, this setup provides an unambiguous oracle for retrieval performance.

% Append/Modify the paragraph starting with "Because questions were generated..."
Because questions were generated from known source passages, this setup provides an unambiguous oracle for retrieval performance. However, we acknowledge that in a large historical corpus, multiple documents may validly support the same answer or provide alternative perspectives. Consequently, our retrieval metrics should be interpreted as a conservative lower bound; they specifically measure the system's ability to recover the exact evidence used to construct the question, even though other relevant documents may exist within the top-$K$ results. To account for this strict exact-match limitation, we provide an extended analysis of ``day-tolerance recall'' and temporal near-misses in Appendix~\ref{sec:appendix_error_analysis}.

Table~\ref{tab:retrieval_results_improved} reports Recall@K, Accuracy@K (i.e the proportion of questions for which \emph{all} gold documents appear in the top-$K$), and MRR@K for the main experimental configurations.

Single-hop retrieval is highly effective: the embedding retriever alone achieves Recall@3 = 67.8, and reranking provides modest gains. This indicates that for questions relying on a single chunk, dense embeddings capture most of the necessary signal.

Multi-hop retrieval is more challenging. Cross-newspaper retrieval achieves low recall with embeddings alone (Recall@3 = 14.3). Cross-domain retrieval (newspaper–debate) performs better (embedding Recall@3 = 47.8). In general cross-domain retrieval performs way better than cross-newspaper retrieval. This indicates strong corpus-bias effect when retrieval is performed on the full database (debates + newspapers): the retriever frequently returns debate chunks even for newspaper–newspaper queries. Concretely, for $K=3$ an average 80.6\% of retrieved documents belonged to the parliamentary debates corpus; this proportion increases with $K$ (83.6\% for $K=5$, 86.8\% for $K=10$). Restricting retrieval to the newspaper-only sub-dataset markedly improves cross-newspaper performance, confirming that source-type priors and corpus-aware retrieval are crucial for cross-source tasks.

A non-obvious finding is that the effect of reranking is \emph{not uniformly positive}. While reranking often improves Recall@10 and MRR@10, it can \emph{reduce} top-$K$ metrics and Accuracy@3 in some configurations. For example, in the newspaper–debate setting embedding-only Recall@3 = 47.8 drops to 45.5 after reranking This behavior likely stems from mismatches between reranker training objectives, domain differences, and corpus biases.

In summary, high single-hop accuracy suggests that answer synthesis, rather than retrieval, is the main bottleneck for simple questions. In contrast, low multi-hop recall and the occasional negative impact of reranking are key limiting factors for end-to-end RAG performance. A detailed breakdown of these retrieval failures—specifically highlighting how the retriever exhibits strong source-type confusion—is provided in Appendix~\ref{sec:appendix_error_analysis}.

\begin{table*}[ht]
\centering
\scriptsize
\setlength{\tabcolsep}{3pt}
\begin{tabular}{l c c c l c c c c c c c c c}
\toprule
\textbf{Category} & \textbf{Queries} & \textbf{Quest. Types} & \textbf{Sources} & \textbf{Method} &
\multicolumn{3}{c}{\textbf{Recall}} &
\multicolumn{3}{c}{\textbf{Accuracy}} &
\multicolumn{3}{c}{\textbf{MRR}} \\
\cmidrule(lr){6-8} \cmidrule(lr){9-11} \cmidrule(lr){12-14}
& & & & & @3 & @5 & @10 & @3 & @5 & @10 & @3 & @5 & @10 \\
\midrule

%-------------------- SINGLE-HOP --------------------%
\rowcolor{gray!10}
\multicolumn{14}{l}{\textbf{Single-hop retrieval}} \\
\midrule
Single-hop (Intra-source) & \nSHqueries & -- & -- & Embedding & 67.8 & 71.6 & 76.4 & 67.8 & 71.6 & 76.4 & 60.6 & 61.5 & 62.1 \\[2pt]
 &  & & & Reranking & 70.6 & 73.1 & 76.4 & 70.6 & 73.1 & 76.4 & 63.1 & 63.7 & 64.1 \\

%-------------------- MULTI-HOP --------------------%
\rowcolor{gray!10}
\multicolumn{14}{l}{\textbf{Multi-hop retrieval}} \\
\midrule

Cross-newspaper Retrieval & 314 & F, C & Gau., Int., Deb. & Embedding & 14.3 & 19.9 & 29.3 & 1.3 & 4.8 & 12.1 & 10.3 & 11.6 & 12.8 \\
& & & & Reranking & 10.4 & 19.0 & 48.4 & 3.5 & 7.0 & 25.2 & 6.2 & 8.1 & 11.9 \\[2pt]

\cmidrule(l){1-14}

Newspaper–Debate Retrieval & 571 & F, C, B & Gau., Int., Deb. & Embedding & 47.8 & 57.4 & 68.6 & 23.3 & 36.3 & 51.5 & 34.6 & 36.8 & 38.3 \\
& & & & Reranking & 45.5 & 55.3 & 74.3 & 17.5 & 28.2 & 57.4 & 35.0 & 37.2 & 40.0 \\[2pt]

\cmidrule(l){1-14}

All Multi-hop (Aggregated) & \nMHqueries & F, C, B & Gau., Int., Deb. & Embedding & 35.9 & 44.1 & 54.6 & 15.5 & 25.1 & 37.5 & 26.0 & 27.8 & 29.2 \\
& & & & Reranking & 33.0 & 42.4 & 65.1 & 12.5 & 20.6 & 56.0 & 24.8 & 27.0 & 29.0 \\[2pt]

%-------------------- ALL --------------------%
\rowcolor{gray!10}
\multicolumn{14}{l}{\textbf{Overall results}} \\
\midrule
All Questions (Single + Multi-hop) & \nTotalQuestions & -- & -- & Embedding & 52.0 & 57.9 & 65.6 & 41.8 & 48.5 & 57.1 & 43.4 & 44.8 & 45.8\\
& & &  & Reranking & 51.9 & 57.9 & 70.8 & 41.8 & 47.0 & 61.3 & 44.1 & 45.4 & 47.1\\[2pt]

\bottomrule
\end{tabular}
\caption{Retrieval performance across query types using \texttt{embed-V4.0} embeddings and \texttt{cohere-rerank-multilingual-v3.0}. Single-hop queries retrieve from one source; multi-hop queries span multiple sources (cross-newspaper within press, or cross-domain between press and debates). Results shown for embedding-based and reranking methods.}
\begin{flushleft}
\footnotesize
Question types: \textbf{F} = Follow-up, \textbf{C} = Comparative, \textbf{B} = Bridge entity.  
Sources: \textbf{Gau.} = \textit{Le Gaulois}, \textbf{Int.} = \textit{L’Intransigeant}, \textbf{Deb.} = \textit{Les Débats parlementaires}.
\end{flushleft}
\label{tab:retrieval_results_improved}
\end{table*}

\subsection{LLM-as-a-Judge Evaluation}

In this section, we evaluate the alignment between human judgments and LLM-as-a-judge evaluations. A subset of $50$ questions, together with their RAG-generated answers, was annotated by our domain expert. Our objective is to assess how closely an LLM-as-a-judge can replicate these human 
evaluations. 

We made multiple iterations over the questions to ensure their relevance and to check for potential hallucinations. RAG answers were carefully read against the source document and annotated as correct, incorrect, or partial. We report the \textit{alignment rate}, defined as the percentage of cases in which the human and the LLM agree. We consider two judging modes:\\

\noindent\textbf{Answer-reference mode :} in which the LLM-as-a-judge is given the question, the RAG-generated answer (based on retrieved documents), and the reference answer produced during the question-generation phase. The LLM is asked to assess whether the RAG answer aligns with the reference answer.\\

\noindent\textbf{Source-oracle mode :} in which the LLM-as-a-judge is given the question, the RAG-generated answer, and the gold documents used to generate the question. The LLM is asked to determine whether the RAG answer is consistent with these source documents.

The first mode requires an intermediate step to generate a reference answer, whereas the second relies directly on the gold documents. We compare both the alignment between human and LLM judgments, and the agreement between the two LLM judging modes.

We experimented with different prompt configurations for the LLM judge, using two models: a base model (\texttt{Command-A}) and a reasoning-enhanced variant. For prompts, we compared (i) a standard formulation allowing partial answers and (ii) a constrained formulation with only two labels (correct or incorrect), in which partially correct human annotations were treated as incorrect. We also evaluated alignment in cases where not all gold documents were successfully retrieved.

The detailed alignment analysis is provided in Table~\ref{tab:confusion_judges}. In the \textit{Source-oracle} setting—where the judge has access to the original chunks and the corresponding LLM-generated answers—the alignment is perfect for instances where RAG outputs are deemed correct by human evaluators. Across the 20 questions labeled as correct by humans, the LLM judge reached full agreement. Conversely, for incorrect answers, the RAG system exhibits markedly lower alignment.

When using a high-capacity model such as Cohere Command-R, the LLM judge appears highly reliable for answers it classifies as correct. Restricting the analysis to RAG systems using such high-quality models further strengthens this observation.After filtering the dataset to retain only instances where RAG answers were validated as correct, 530 questions remain, distributed across the \textit{Follow-up} (300), \textit{Comparative} (121), and \textit{Bridge Entity} (109) categories.

\begin{table*}[ht]
\centering
\small
\setlength{\tabcolsep}{4pt}
\begin{tabular}{
  lccc@{\hskip 1.8em}lccc
}
\toprule
\multicolumn{4}{c}{\textbf{Answer-reference}} &
\multicolumn{4}{c}{\textbf{Source-oracle}} \\
\cmidrule(lr){1-4} \cmidrule(lr){5-8}
& \multicolumn{3}{c}{\textbf{LLM Annotation}} &
& \multicolumn{3}{c}{\textbf{LLM Annotation}} \\
\cmidrule(lr){2-4} \cmidrule(lr){6-8}
\textbf{Human Annotation} & \textbf{Correct} & \textbf{Incorrect} & \textbf{Total} &
\textbf{Human Annotation} & \textbf{Correct} & \textbf{Incorrect} & \textbf{Total} \\
\midrule
Correct   & \textbf{10} & 10 & 20 & Correct   & \textbf{20} & 0  & 20 \\
Incorrect & 6  & \textbf{24} & 30 & Incorrect & 16 & \textbf{14} & 30 \\
\midrule
\textbf{Total} & 16 & 34 & 50 & \textbf{Total} & 36 & 14 & 50 \\
\bottomrule
\end{tabular}
\caption{
Confusion matrices comparing LLM-based judges with human annotations. Two settings are considered: (1) \textit{Answer-reference}, which assesses answers against a reference answer, and (2) \textit{Source-oracle}, which evaluates answers based on the source documents. %Rows correspond to \textbf{human annotations}, while columns correspond to \textbf{LLM annotations}.  
}
\label{tab:confusion_judges}
\end{table*}

\subsection{RAG Answer Evaluation}

% Replace the opening sentence of Section 5.4:
After filtering for questions where the RAG answers were correctly evaluated by the LLM-as-a-Judge, we perform a full evaluation on this validated high-confidence subset. While we recognize that restricting the analysis to instances where the judge and human expert reach consensus introduces a selection bias toward clearer examples, this methodology was chosen to ensure that the reported RAG performance reflects model reasoning capabilities rather than evaluator noise or "judge hallucinations." In this setup, we do not operate in oracle mode, allowing the retriever to potentially return incorrect documents. We use a configuration with $K=3$ retrieved documents and no reranking, as reranking was found to negatively affect performance on the multi-hop question dataset. The RAG system is prompted to answer each question based on the retrieved documents, and the resulting answers are evaluated by the LLM-as-a-Judge. This evaluation assesses how effectively each RAG pipeline retrieves and synthesizes information from the source corpus, as reflected in the accuracy scores provided by the LLM evaluator.

Table~\ref{tab:llm_eval_clean} reports results for a set of models differing in architecture. \texttt{gpt-oss-20b} attains the highest accuracy on both single-hop (58.85\%) and multi-hop (18.86\%) questions. \texttt{Llama-3.3-70B-Instruct} (53.53\% single-hop, 17.92\% multi-hop) and \texttt{gpt-4o-mini} (52.72\% single-hop, 13.77\% multi-hop) follow in performance. \texttt{Llama-3.3-3b-Instruct} shows the lowest performance on both tasks, with 24.10\% on single-hop and 3.2\% on multi-hop questions.

The gap between single- and multi-hop performance underscores the persistent challenge of compositional reasoning in retrieval-augmented generation. While single-hop accuracy varies significantly (ranging from 24.10\% to 58.85\%), scores drop substantially on multi-hop questions for all models. This pattern suggests that retrieval quality alone is insufficient: the answering step requires analyzing multiple and potentially long chunks. Effective integration and reasoning across multiple contexts remain critical limiting factors.

%% file: content/6_conclusion.tex
\section{Conclusion and Future Works}

We introduced a French-language dataset of \nSHqueries ~\ single-hop and \nMHqueries ~\ multi-hop questions from French Third Republic parliamentary debates and newspapers, designed to advance RAG system evaluation for historical research. Our historian-guided methodology ensures questions reflect genuine historical inquiry.

While single-hop retrieval performs well, multi-hop reasoning and cross-source synthesis remain challenging. Our experiments used Cohere API models for their benchmark performance. Future work will compare alternative models across question generation, retrieval, reranking, and LLM evaluation; expand the dataset; and refine RAG architectures to improve cross-source reasoning and assess model robustness.

\begin{comment}
In this paper, we introduced a novel French-language dataset comprising 897 single-hop and 885 multi-hop questions derived from parliamentary debates and newspapers of the French Third Republic. The dataset is designed to advance the development and evaluation of Retrieval-Augmented Generation (RAG) systems in the context of historical research. Our historian-guided methodology ensures that the generated questions are both accurate and meaningful, reflecting the depth and complexity of genuine historical inquiry.

While single-hop retrieval achieves strong performance, multi-hop reasoning and cross-source synthesis continue to pose significant challenges, highlighting key areas for future investigation. 

 Our initial experiments relied on models from the Cohere API, chosen for their strong performance on various benchmarks. 
 As future work We intend to extend this evaluation by comparing question generation, retrieval, reranking, and LLM-as-a-judge setups across alternative models. We also plan to expand the dataset, diversify question types, and refine RAG architectures to improve cross-source integration and reasoning capabilities to further assess model robustness and adaptability.
\end{comment}

\section{Limitations}

We relied on Cohere API models for question generation, retrieval, and alignment evaluation, which means results may differ with alternative models. In addition, the question generation and manual validation were performed by a single historian; future work would benefit from multiple annotators to assess inter-annotator agreement. Furthermore, the corpus is limited to a single year (1887) and two newspapers, which may not capture the full diversity of the period. Finally, our LLM-as-a-judge alignment was validated on a small subset of 50 questions. Furthermore, our final RAG performance analysis is restricted to the subset where human and LLM judgments aligned; this likely overestimates absolute performance by excluding ambiguous or high-disagreement cases. Future research should investigate judge performance specifically on these "hard" cases to better define the boundaries of automated evaluation in historical RAG systems.

\section{Ethical Considerations}

The historical documents used in our corpus are in the public domain, sourced from the Bibliothèque nationale de France. These texts contain viewpoints and biases from their period (1887), which the dataset is designed to capture for research purposes, not to endorse. To mitigate the risks associated with historically biased or misleading interpretations, we adopted a historian-in-the-loop methodology for question validation.

%% file: content/7_appendix_error_analysis.tex
\clearpage
\onecolumn
\appendix

\section{Retrieval Error Analysis}
\label{sec:appendix_error_analysis}

We complement the retrieval evaluation of Section~5 by analyzing error patterns along two dimensions: (1)~the proximity of failed single-hop retrievals to the gold document, and (2)~the relationship between multi-hop question types, source-type confusion, and retrieval performance.

\subsection{Single-Hop: Proximity of Retrieved Documents}

\begin{figure}[ht]
\centering
\includegraphics[width=\azer\linewidth]{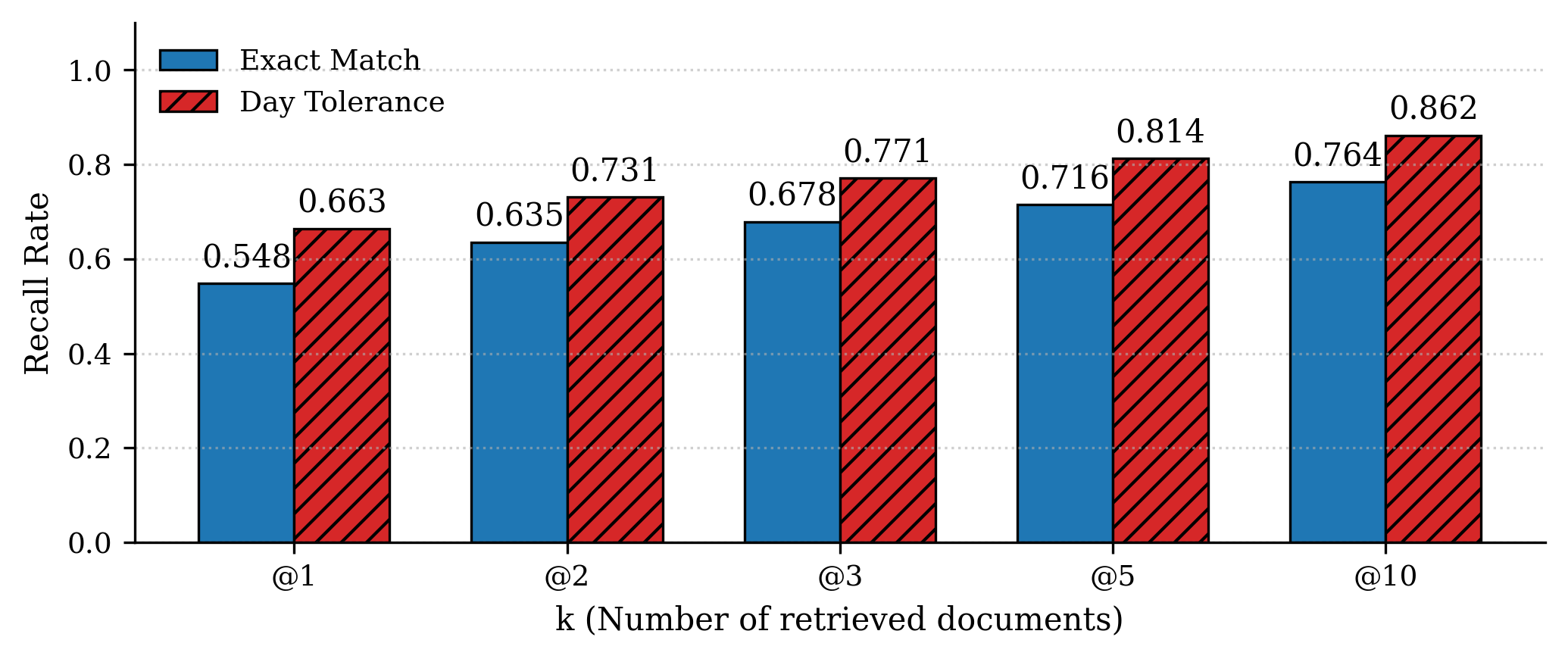}
\caption{Exact-match recall vs.\ day-tolerance recall at different $k$ values for single-hop questions. Day-tolerance considers a retrieval successful if any retrieved document shares the same publication date as the gold document.}
\label{fig:day_tolerance_recall}
\end{figure}

Standard recall treats retrieval as binary: either the exact gold chunk is found or it is not. However, many apparent failures retrieve a document from the \emph{same publication date} as the gold chunk, indicating that the retriever captures temporal context but selects an adjacent passage. Figure~\ref{fig:day_tolerance_recall} compares exact-match recall with \emph{day-tolerance recall}, where a retrieval counts as successful if any top-$k$ document shares the publication date of the gold document.

The gap between exact and day-tolerance recall is consistent across all $k$ values, showing that a substantial fraction of errors are near-misses rather than complete failures. Table~\ref{tab:error_breakdown} provides a detailed breakdown: while exact-match recall plateaus at 76.4\% for top-10, same-day retrieval reaches a notably higher rate, leaving only a minority of questions as complete misses where no temporally related document appears in the top-10.

\begin{table}[ht]
\centering
\small
\begin{tabular}{lccc}
\toprule
\textbf{Criterion} & \textbf{@1} & \textbf{@5} & \textbf{@10} \\
\midrule
Exact match     & 54.8\% & 71.6\% & 76.4\% \\
Same day        & 66.3\% & 81.4\% & 86.2\% \\
Complete miss   & 33.7\% & 18.6\% & 13.8\% \\
\bottomrule
\end{tabular}
\caption{Single-hop retrieval error breakdown (897 questions). \emph{Exact match}: gold chunk found; \emph{Same day}: a document from the same publication date found; \emph{Complete miss}: no same-day document in top-$k$.}
\label{tab:error_breakdown}
\end{table}

\begin{figure}[ht]
\centering
\includegraphics[width=\azer\linewidth]{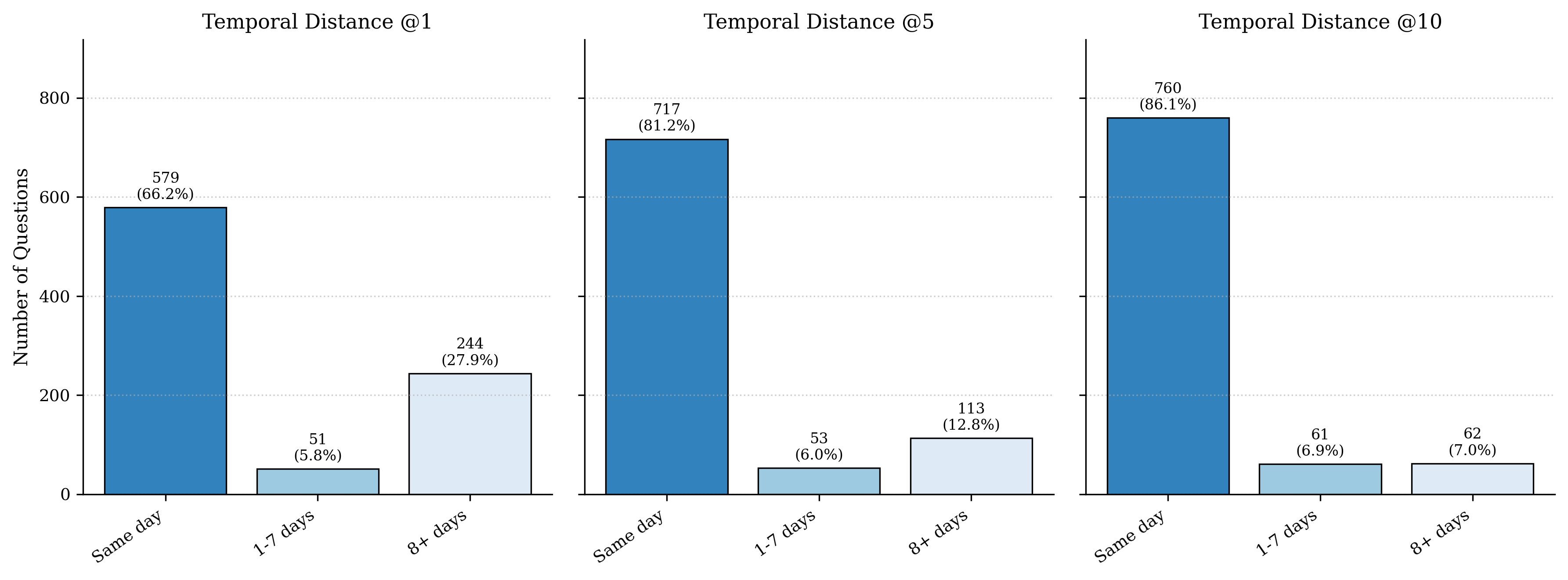}
\caption{Distribution of minimum temporal distance (in days) between the closest retrieved document and the gold document, at $k \in \{1, 5, 10\}$.}
\label{fig:temporal_distance}
\end{figure}

\paragraph{Temporal and chunk distance.}
Figure~\ref{fig:temporal_distance} shows the distribution of the minimum temporal distance between retrieved and gold documents. At $k=10$, the majority of questions have a same-day document in the top results, and the remaining are equally distributed between a 1--7 day window and an 8+ day window.

\begin{figure}[ht]
\centering
\includegraphics[width=\azer\linewidth]{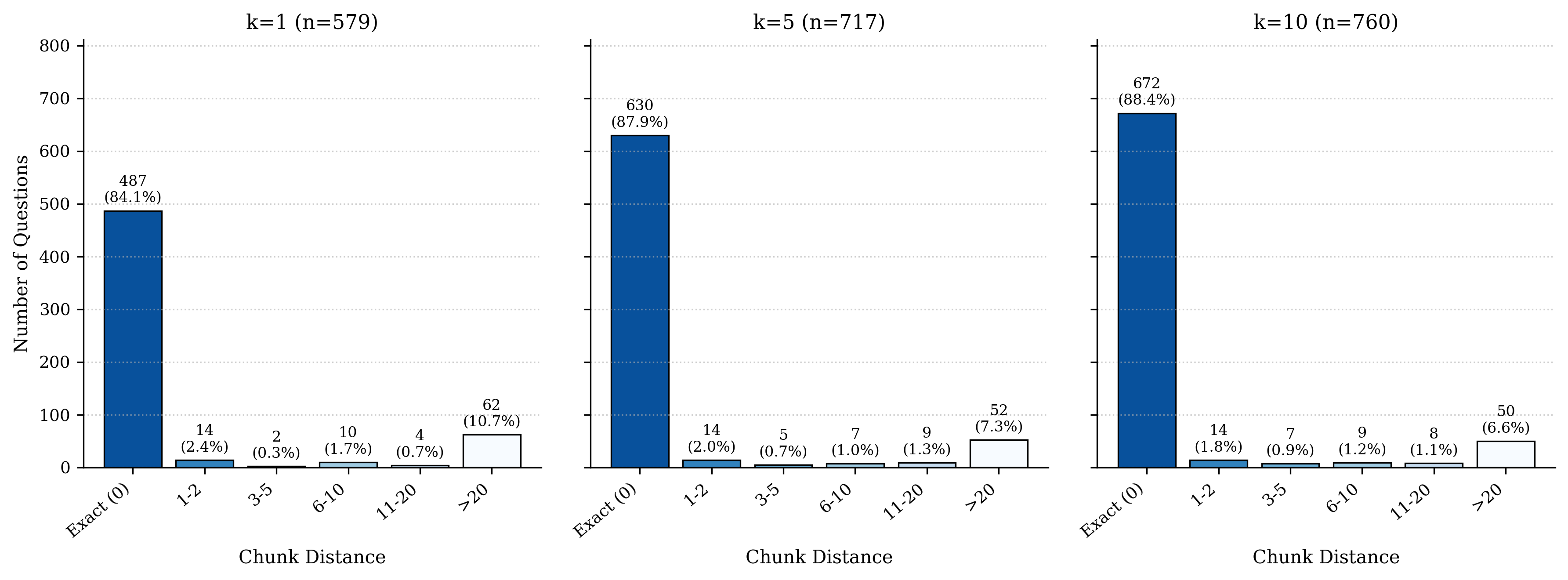}
\caption{Chunk distance distribution for same-day retrieved documents at $k \in \{1, 5, 10\}$. Most near-misses retrieve exact or adjacent chunk.}
\label{fig:chunk_distance}
\end{figure}

For same-day retrievals that miss the exact chunk, Figure~\ref{fig:chunk_distance} reports the chunk distance distribution. The majority of near-misses fall within a long distance from the actual chunk, suggesting that the retriever model is, most of the time, either able to find the correct chunk or misses it by a lot.

\clearpage

\subsection{Multi-Hop: Question Type and Source Confusion}

\begin{figure}[ht]
\centering
\includegraphics[width=\azer\linewidth]{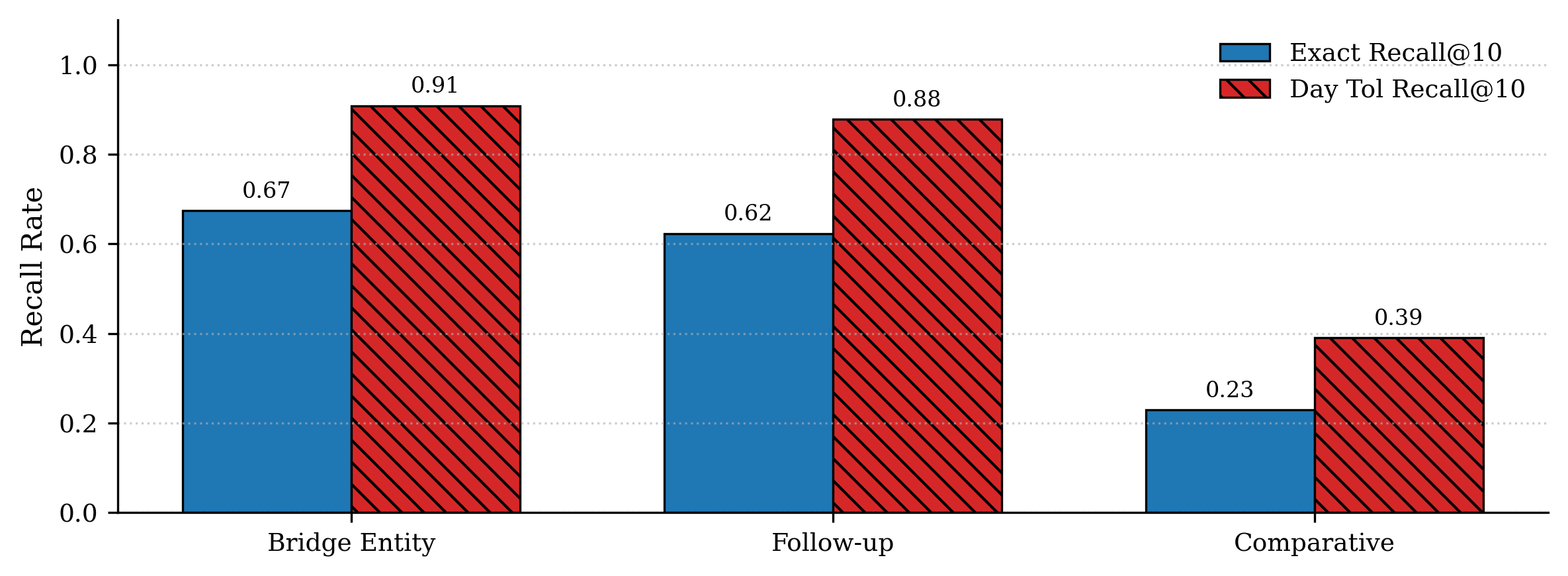}
\caption{Exact recall@10 vs.\ day-tolerance recall@10 across multi-hop question types.}
\label{fig:multihop_by_type}
\end{figure}

Figure~\ref{fig:multihop_by_type} breaks down retrieval performance by multi-hop question category (Exact Recall@10 vs.\ Day-Tolerance Recall@10). The gap between the two metrics varies across types, revealing that certain categories are disproportionately affected by near-miss chunk selection.

\begin{figure}[ht]
\centering
\includegraphics[width=\azer\linewidth]{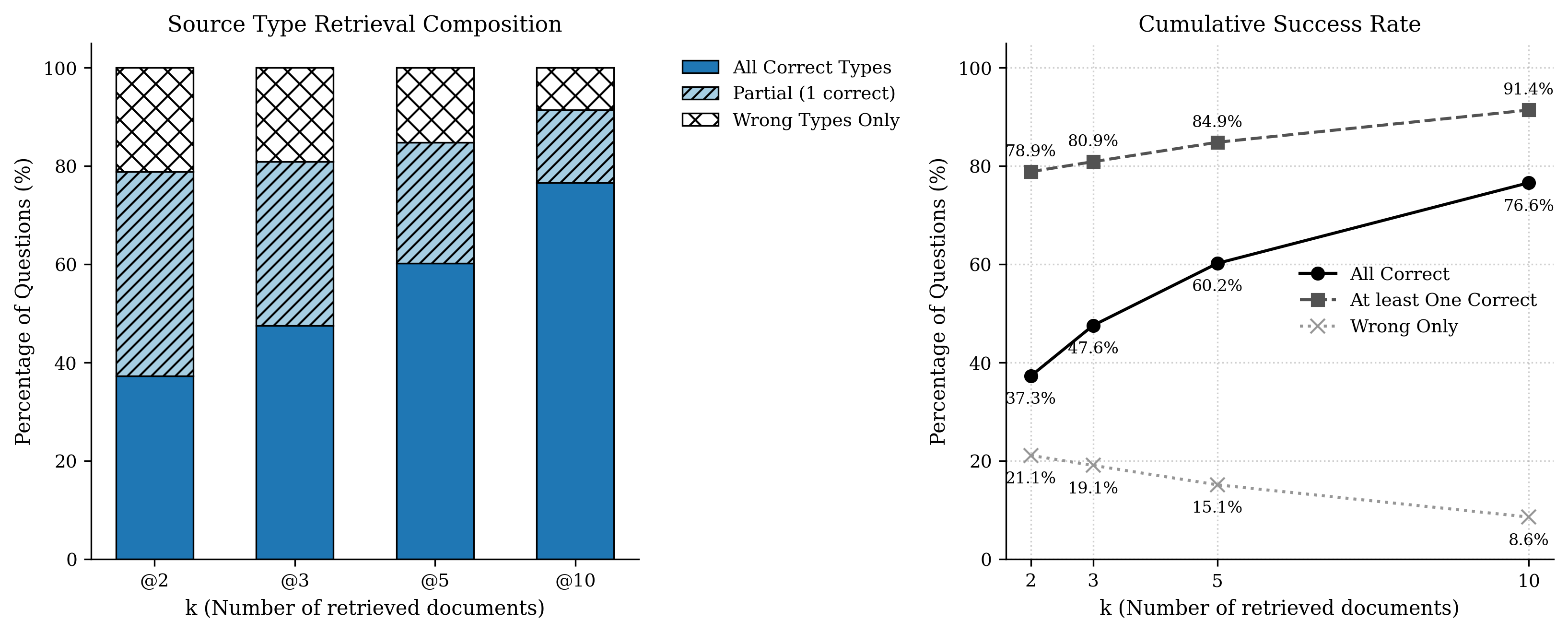}
\caption{Source-type retrieval status at different $k$ values (left: stacked proportions; right: trend lines). \emph{All Correct}: all expected source types present; \emph{Partial}: at least one correct type; \emph{Wrong Only}: no expected type found.}
\label{fig:source_type_perf}
\end{figure}

\paragraph{Source-type retrieval across $k$.}
Figure~\ref{fig:source_type_perf} tracks how the proportion of questions with correct source types evolves across $k$ values. At $k{=}2$, 37.3\% of questions retrieve all expected source types, increasing to 76.6\% at $k{=}10$. Even at $k{=}10$, 8.6\% of questions return only wrong source types. The ``wrong only'' category decreases steadily with $k$, but never vanishes, indicating a persistent structural mismatch for a subset of queries.

\begin{figure}[ht]
\centering
\includegraphics[width=\azer\linewidth]{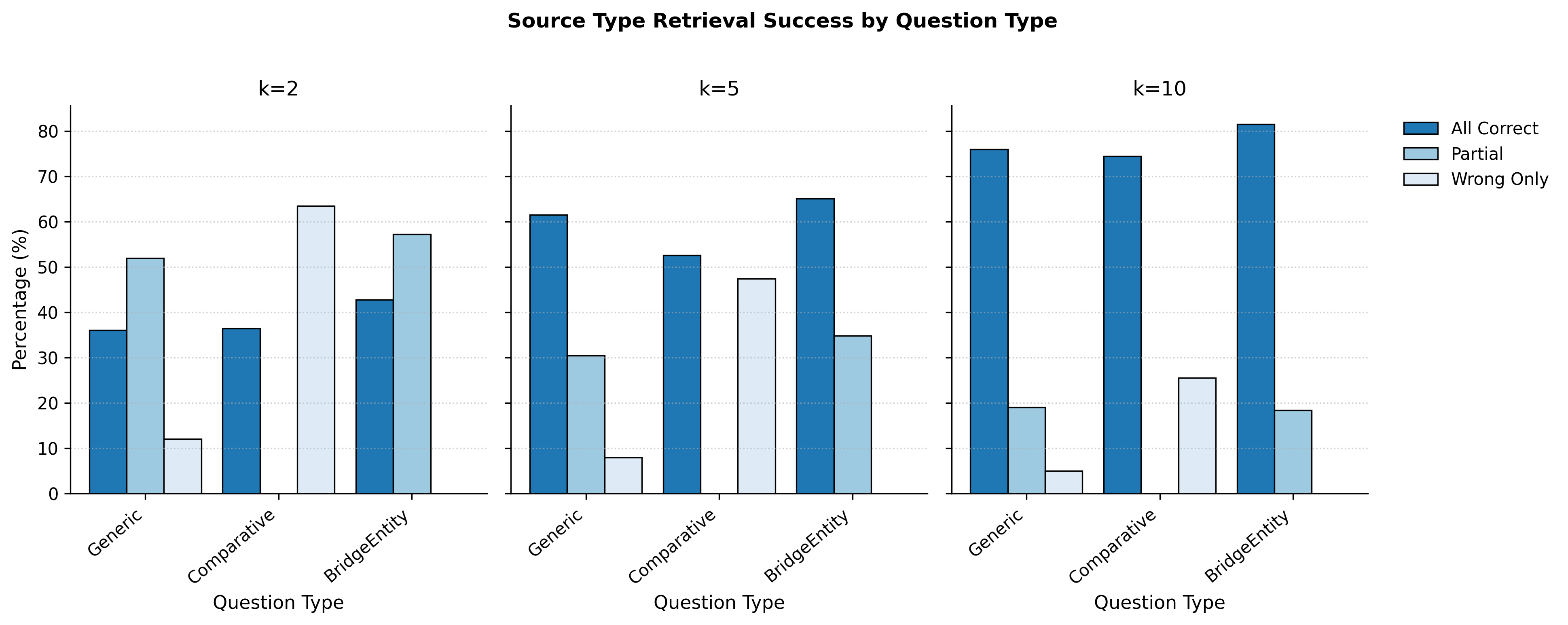}
\caption{Source-type retrieval status (\% All Correct / Partial / Wrong Only) broken down by question type at $k \in \{1, 5, 10\}$.}
\label{fig:source_type_by_qtype}
\end{figure}

\paragraph{Source-type confusion by question type.}
Figure~\ref{fig:source_type_by_qtype} breaks down the source-type retrieval status by question category. Comparative questions show the highest ``wrong only'' rate (25.5\% at $k{=}10$), as they require cross-newspaper retrieval which is most affected by the debate corpus bias. Bridge Entity questions never have ``wrong only'' errors---their debate--newspaper structure naturally aligns with the retriever's tendency to return debate chunks. Generic (Follow-up) questions fall in between.

\begin{figure}[ht]
\centering
\includegraphics[width=\azer\linewidth]{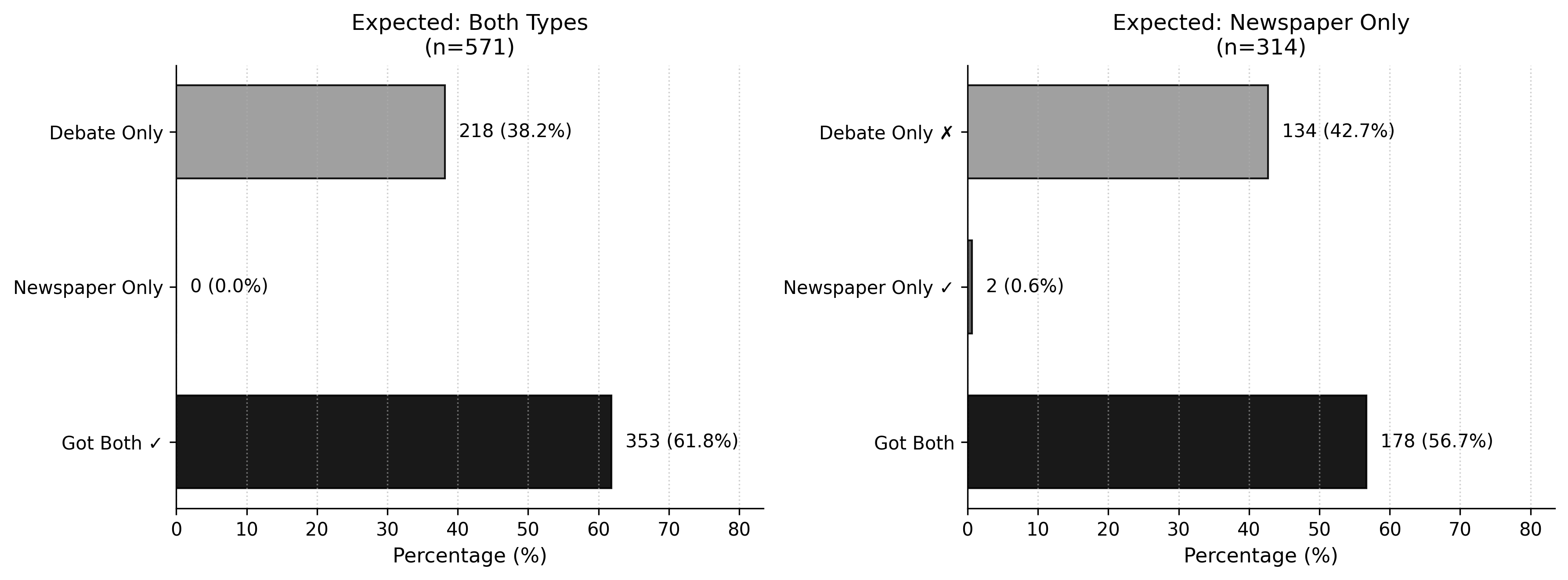}
\caption{Source-type confusion analysis (top-5 retrieved). Left: questions expecting both newspaper and debate sources. Right: questions expecting only newspaper sources. The retriever overwhelmingly returns debate chunks even when newspapers are expected.}
\label{fig:source_confusion}
\end{figure}

\paragraph{Source-type confusion analysis.}
Figure~\ref{fig:source_confusion}
%and Table~\ref{tab:source_confusion} 
details the source-type confusion patterns at $k{=}5$ and compare expected vs.\ retrieved source types. For questions requiring both newspaper and debate documents, the retriever consistently over-represents debate chunks---consistent with the corpus bias reported in Section~5 (80.6\% debate chunks at $K{=}3$). Questions expecting exclusively cross-newspaper sources are most affected: the retriever frequently returns debate passages instead of the required newspaper pairs. This asymmetry confirms that source-aware retrieval strategies, such as forced multi-collection distribution or classification-based routing, are essential for cross-source multi-hop question answering on heterogeneous historical corpora.

% \begin{table}[ht]
% \centering
% \small
% \caption{Source-type confusion analysis (top-5 retrieved, 885 multi-hop questions). For each category of expected sources, we report the proportion of questions where the retrieved set contains both types, the correct type only, or only the wrong type.}
% \label{tab:source_confusion}
% \begin{tabular}{lccccc}
% \toprule
% \textbf{Expected} & \textbf{$n$} & \textbf{Got Both} & \textbf{Correct Only} & \textbf{Wrong Only} & \textbf{None} \\
% \midrule
% Both (mixed)      & 571 & 61.8\% & ---    & 38.2\% & 0\% \\
% Newspaper only    & 314 & 56.7\% & 0.6\%  & 42.7\% & 0\% \\
% \bottomrule
% \end{tabular}
% \end{table}

%\clearpage

\section{Prompt Specifications for Question Generation}
\label{sec:appendix_prompts}

This appendix documents the system prompts used for multi-hop question generation in the HistoriQA-ThirdRepublic dataset. All prompts were written in French to match the language of the source corpora. We present below one representative prompt in full (Follow-up: Debate--Newspaper), followed by a summary of the four remaining prompts. The complete set of prompts is available in the accompanying \href{https://github.com/atomegoyan/historiqa-thirdrepublic}{repository}.\\

\noindent \textbf{Type:} System message \\
\textbf{Purpose:} Generate multi-hop follow-up questions requiring cross-referencing between a parliamentary debate passage and a newspaper article reporting on the same debate.

\begin{promptbox}[System Instructions]
Tu es un assistant spécialisé dans la génération de questions
multihop à partir de deux sources :
- Extrait A : un passage d'un débat parlementaire de la
  Troisième République
- Extrait B : un article de presse (par exemple
  L'Intransigeant, rubrique "La Chambre") rendant compte
  de ce même débat.

### Règles :
- Chaque question doit IMPÉRATIVEMENT nécessiter de croiser
  les deux extraits.
- Si aucune question multihop pertinente ne peut être
  produite, répondre exactement "AUCUNE QUESTION".
- Les questions doivent être compréhensibles sans les
  extraits : mentionner la date (si présente), le sujet
  central (loi, budget, événement) et les acteurs
  principaux (députés, orateurs).
- Évite toute formulation vague du type "selon l'article
  de presse".
- Mets en avant les enjeux politiques, les arguments
  centraux ou la perception publique, pas des détails
  accessoires.
- Varie les types de raisonnement : factuel, chronologique,
  causal, comparatif, critique/inférentiel.
- Chaque question doit être suivie de sa réponse, des
  passages justificatifs, du type de raisonnement et du
  niveau de difficulté.

### Format attendu :
[
  {
    "question": "...",
    "answer": "...",
    "supporting_passages": {
      "debate": "...",
      "journal": "..."
    },
    "reasoning_type": "...",
    "difficulty": "..."
  },
  ...
]

---

### Exemples attendus :

Exemple 1 :
[
  {
    "question": "En 1887, comment M. Pichon justifie-t-il
      la suppression du budget des cultes dans son discours
      à la Chambre, et quelle réaction la presse appelle-
      t-elle de la part des électeurs face au vote de la
      majorité ?",
    "answer": "M. Pichon soutient que ni l'Église ni
      l'État n'exécutent le Concordat, ce qui invalide son
      maintien. La presse espère que les électeurs
      imposeront à leurs mandataires une ligne de conduite
      différente, marquant un mécontentement face au vote
      de la majorité.",
    "supporting_passages": {
      "debate": "Je vous demande, messieurs, si vous êtes
        sûrs d'être liés par le Concordat.",
      "journal": "Le pays jugera, et il faut espérer que
        les électeurs sauront imposer désormais à leurs
        mandataires une ligne de conduite conforme à leurs
        desiderata."
    },
    "reasoning_type": "critique/inférentiel",
    "difficulty": "moyen"
  }
]

Exemple 2 :
[
  {
    "question": "Quels arguments M. Pichon avance-t-il à
      la Chambre pour supprimer le budget des cultes, et
      comment la presse décrit-elle la majorité
      parlementaire qui a rejeté cette proposition ?",
    "answer": "M. Pichon affirme que ni l'Église ni l'État
      n'exécutent le Concordat. La presse décrit la
      majorité comme une alliance opportuniste entre les
      amis de Ferry et de Freppel, représentant environ
      300 voix.",
    "supporting_passages": {
      "debate": "Je dis que ni l'Eglise ni l'État
        n'exécutent le Concordat.",
      "journal": "Une majorité de 300 voix recrutée parmi
        les amis de MM. Ferry et Freppel a décidé qu'il
        était indispensable de rétribuer grassement les
        prêtres, aux dépens des contribuables."
    },
    "reasoning_type": "comparatif",
    "difficulty": "difficile"
  }
]
\end{promptbox}

\subsubsection*{User Message}

The user message contains the two text excerpts to be analysed, provided as \textit{Extrait~A} (parliamentary debate passage) and \textit{Extrait~B} (newspaper article).

\begin{promptbox}[System Instructions (English Translation)]
You are an assistant specialized in generating multihop questions from two sources:
- Excerpt A: a passage from a parliamentary debate of the Third Republic
- Excerpt B: a newspaper article (e.g., L'Intransigeant, "La Chambre" column) reporting on this same debate.

### Rules:
- Each question MUST IMPERATIVELY require cross-referencing the two excerpts.
- If no relevant multihop question can be produced, answer exactly "AUCUNE QUESTION" (NO QUESTION).
- The questions must be understandable without the excerpts: mention the date (if present), the central subject (law, budget, event) and the main actors (deputies, speakers).
- Avoid any vague formulation like "according to the newspaper article".
- Highlight political stakes, central arguments or public perception, not incidental details.
- Vary the reasoning types: factual, chronological, causal, comparative, critical/inferential.
- Each question must be followed by its answer, the supporting passages, the reasoning type, and the difficulty level.

### Expected format:
[
  {
    "question": "...",
    "answer": "...",
    "supporting_passages": {
      "debate": "...",
      "journal": "..."
    },
    "reasoning_type": "...",
    "difficulty": "..."
  },
  ...
]

---

### Expected examples:

Example 1:
[
  {
    "question": "In 1887, how does Mr. Pichon justify the suppression of the budget of cults in his speech to the Chamber, and what reaction does the press call for from the voters in response to the majority's vote?",
    "answer": "Mr. Pichon maintains that neither the Church nor the State is executing the Concordat, which invalidates its continuation. The press hopes that voters will impose a different line of conduct on their representatives, marking dissatisfaction with the majority's vote.",
    "supporting_passages": {
      "debate": "I ask you, gentlemen, if you are sure you are bound by the Concordat.",
      "journal": "The country will judge, and it is to be hoped that voters will henceforth know how to impose on their representatives a line of conduct in accordance with their wishes."
    },
    "reasoning_type": "critical/inferential",
    "difficulty": "medium"
  }
]

Example 2:
[
  {
    "question": "What arguments does Mr. Pichon put forward to the Chamber to suppress the budget of cults, and how does the press describe the parliamentary majority that rejected this proposal?",
    "answer": "Mr. Pichon asserts that neither the Church nor the State is executing the Concordat. The press describes the majority as an opportunistic alliance between the friends of Ferry and Freppel, representing about 300 votes.",
    "supporting_passages": {
      "debate": "I say that neither the Church nor the State executes the Concordat.",
      "journal": "A majority of 300 votes recruited among the friends of Messrs. Ferry and Freppel decided that it was indispensable to pay the priests handsomely, at the expense of the taxpayers."
    },
    "reasoning_type": "comparative",
    "difficulty": "hard"
  }
]
\end{promptbox}

%% file: 0_main.bbl
\begin{thebibliography}{21}
\expandafter\ifx\csname natexlab\endcsname\relax\def\natexlab#1{#1}\fi

\bibitem[{Chang et~al.(2025)Chang, Jiang, Rakesh, Pan, Yeh, Wang, Hu, Xu,
  Zheng, Das, and
  Zou}]{Chang_Jiang_Rakesh_Pan_Yeh_Wang_Hu_Xu_Zheng_Das_et_al._2025}
Chia-Yuan Chang, Zhimeng Jiang, Vineeth Rakesh, Menghai Pan, Chin-Chia~Michael
  Yeh, Guanchu Wang, Mingzhi Hu, Zhichao Xu, Yan Zheng, Mahashweta Das, and
  Na~Zou. 2025.
\newblock \href {https://doi.org/10.18653/v1/2025.acl-long.131} {Main-rag:
  Multi-agent filtering retrieval-augmented generation}.
\newblock In \emph{Proceedings of the 63rd Annual Meeting of the Association
  for Computational Linguistics (Volume 1: Long Papers)}, page 2607–2622,
  Vienna, Austria. Association for Computational Linguistics.

\bibitem[{Grattafiori et~al.(2024)Grattafiori, Dubey, Jauhri, Pandey, Kadian,
  Al-Dahle, Letman, Mathur, Schelten, Vaughan, Yang, Fan, Goyal, Hartshorn,
  Yang, Mitra, Sravankumar, Korenev, Hinsvark, Rao, Zhang, and
  Rodriguez}]{llama3}
Aaron Grattafiori, Abhimanyu Dubey, Abhinav Jauhri, Abhinav Pandey, Abhishek
  Kadian, Ahmad Al-Dahle, Aiesha Letman, Akhil Mathur, Alan Schelten, Alex
  Vaughan, Amy Yang, Angela Fan, Anirudh Goyal, Anthony Hartshorn, Aobo Yang,
  Archi Mitra, Archie Sravankumar, Artem Korenev, Arthur Hinsvark, Arun Rao,
  Aston Zhang, and Aurelien Rodriguez. 2024.
\newblock \href {http://arxiv.org/abs/2407.21783} {The llama 3 herd of models}.

\bibitem[{Guo et~al.(2024)Guo, Guo, Su, Yang, Zhu, Li, Qiu, and
  Liu}]{guo2024biaslargelanguagemodels}
Yufei Guo, Muzhe Guo, Juntao Su, Zhou Yang, Mengqiu Zhu, Hongfei Li, Mengyang
  Qiu, and Shuo~Shuo Liu. 2024.
\newblock \href {http://arxiv.org/abs/2411.10915} {Bias in large language
  models: Origin, evaluation, and mitigation}.

\bibitem[{Hendrycks et~al.(2021)Hendrycks, Burns, Basart, Zou, Mazeika, Song,
  and Steinhardt}]{mmlu}
Dan Hendrycks, Collin Burns, Steven Basart, Andy Zou, Mantas Mazeika, Dawn
  Song, and Jacob Steinhardt. 2021.
\newblock \href {http://arxiv.org/abs/2009.03300} {Measuring massive multitask
  language understanding}.

\bibitem[{Ho et~al.(2020)Ho, Duong~Nguyen, Sugawara, and
  Aizawa}]{ho-etal-2020-constructing}
Xanh Ho, Anh-Khoa Duong~Nguyen, Saku Sugawara, and Akiko Aizawa. 2020.
\newblock \href {https://doi.org/10.18653/v1/2020.coling-main.580}
  {Constructing a multi-hop {QA} dataset for comprehensive evaluation of
  reasoning steps}.
\newblock In \emph{Proceedings of the 28th International Conference on
  Computational Linguistics}, pages 6609--6625, Barcelona, Spain (Online).
  International Committee on Computational Linguistics.

\bibitem[{Hwang et~al.(2024)Hwang, Kim, and Lee}]{Hwang_Kim_Lee_2024}
Seonjeong Hwang, Yunsu Kim, and Gary~Geunbae Lee. 2024.
\newblock \href {https://doi.org/10.48550/ARXIV.2404.00571} {Explainable
  multi-hop question generation: An end-to-end approach without intermediate
  question labeling}.

\bibitem[{Kalifa et~al.(2011)Kalifa, Régnier, and Ève
  Thérenty}]{kalifa_trerenty_2011}
Dominique Kalifa, Philippe Régnier, and Marie Ève Thérenty, editors. 2011.
\newblock \emph{La civilisation du journal : histoire culturelle et littéraire
  de la presse française au XIXe siècle}.
\newblock Nouveau Monde Éditions, Paris.

\bibitem[{Leng et~al.(2024)Leng, Portes, Havens, Zaharia, and
  Carbin}]{Leng_Portes_Havens_Zaharia_Carbin_2024}
Quinn Leng, Jacob Portes, Sam Havens, Matei Zaharia, and Michael Carbin. 2024.
\newblock \href
  {https://www.databricks.com/blog/long-context-rag-performance-llms} {Long
  context rag performance of llms}.

\bibitem[{Lewis et~al.(2020)Lewis, Perez, Piktus, Petroni, Karpukhin, Goyal,
  Küttler, Lewis, Yih, Rocktäschel, Riedel, and
  Kiela}]{Lewis_Perez_Piktus_Petroni_Karpukhin_Goyal_Küttler_Lewis_Yih_Rocktäschel_et_al._2020}
Patrick Lewis, Ethan Perez, Aleksandra Piktus, Fabio Petroni, Vladimir
  Karpukhin, Naman Goyal, Heinrich Küttler, Mike Lewis, Wen-tau Yih, Tim
  Rocktäschel, Sebastian Riedel, and Douwe Kiela. 2020.
\newblock \href
  {https://proceedings.neurips.cc/paper_files/paper/2020/file/6b493230205f780e1bc26945df7481e5-Paper.pdf}
  {Retrieval-augmented generation for knowledge-intensive nlp tasks}.
\newblock In \emph{Advances in Neural Information Processing Systems},
  volume~33, page 9459–9474. Curran Associates, Inc.

\bibitem[{Li et~al.(2025{\natexlab{a}})Li, Zhang, and
  Kong}]{Li_Zhang_Kong_2025}
Maodong Li, Longyin Zhang, and Fang Kong. 2025{\natexlab{a}}.
\newblock \href {https://doi.org/10.18653/v1/2025.findings-acl.526} {Multi-hop
  question generation via dual-perspective keyword guidance}.
\newblock In \emph{Findings of the Association for Computational Linguistics:
  ACL 2025}, page 10096–10112, Vienna, Austria. Association for Computational
  Linguistics.

\bibitem[{Li et~al.(2025{\natexlab{b}})Li, He, Liu, Zhang, Yu, Ye, Zhu, and
  Su}]{Li_He_Liu_Zhang_Yu_Ye_Zhu_Su_2025}
Rui Li, Liyang He, Qi~Liu, Zheng Zhang, Heng Yu, Yuyang Ye, Linbo Zhu, and
  Yu~Su. 2025{\natexlab{b}}.
\newblock \href {https://doi.org/10.18653/v1/2025.acl-long.693} {Unirag:
  Unified query understanding method for retrieval augmented generation}.
\newblock In \emph{Proceedings of the 63rd Annual Meeting of the Association
  for Computational Linguistics (Volume 1: Long Papers)}, page 14163–14178,
  Vienna, Austria. Association for Computational Linguistics.

\bibitem[{Lin et~al.(2024)Lin, Chen, Song, and
  Zhang}]{Lin_Chen_Song_Zhang_2024}
Zefeng Lin, Weidong Chen, Yan Song, and Yongdong Zhang. 2024.
\newblock \href {https://doi.org/10.18653/v1/2024.findings-naacl.236}
  {Prompting few-shot multi-hop question generation via comprehending
  type-aware semantics}.
\newblock In \emph{Findings of the Association for Computational Linguistics:
  NAACL 2024}, page 3730–3740, Mexico City, Mexico. Association for
  Computational Linguistics.

\bibitem[{McCombs and Shaw(1972)}]{McCombs_Shaw_1972}
Maxwell~E. McCombs and Donald~L. Shaw. 1972.
\newblock \href {http://www.jstor.org/stable/2747787} {The agenda-setting
  function of mass media}.
\newblock \emph{The Public Opinion Quarterly}, 36(2):176--187.

\bibitem[{Morel(2024)}]{Morel_2024}
Benjamin Morel. 2024.
\newblock \emph{Le Parlement, temple de la République : de 1789 à nos jours}.
\newblock Passés composés, Paris.

\bibitem[{OpenAI et~al.(2024)OpenAI, Achiam, Adler, Agarwal, Ahmad, Akkaya,
  Aleman, Almeida, Altenschmidt, Altman, Anadkat, Avila, Babuschkin, Balaji,
  Balcom, Baltescu, Bao, Bavarian, Belgum, Bello, Berdine, Bernadett-Shapiro,
  Berner, Bogdonoff, Boiko, Boyd, Brakman, Brockman, and Brooks}]{gpt4}
OpenAI, Josh Achiam, Steven Adler, Sandhini Agarwal, Lama Ahmad, Ilge Akkaya,
  Florencia~Leoni Aleman, Diogo Almeida, Janko Altenschmidt, Sam Altman,
  Shyamal Anadkat, Red Avila, Igor Babuschkin, Suchir Balaji, Valerie Balcom,
  Paul Baltescu, Haiming Bao, Mohammad Bavarian, Jeff Belgum, Irwan Bello, Jake
  Berdine, Gabriel Bernadett-Shapiro, Christopher Berner, Lenny Bogdonoff, Oleg
  Boiko, Madelaine Boyd, Anna-Luisa Brakman, Greg Brockman, and Tim Brooks.
  2024.
\newblock \href {http://arxiv.org/abs/2303.08774} {Gpt-4 technical report}.

\bibitem[{Pellet et~al.(2025)Pellet, Perez, and Puren}]{pellet:hal-05193494}
Aur{\'e}lien Pellet, Julien Perez, and Marie Puren. 2025.
\newblock \href {https://hal.science/hal-05193494} {{{\'E}valuation automatique
  du retour {\`a} la source dans un contexte historique long et bruit{\'e} :
  les d{\'e}bats parlementaires de la Troisi{\`e}me R{\'e}publique fran{\c
  c}aise}}.
\newblock In \emph{{PFIA 2025}}, Dijon, France.

\bibitem[{Phan et~al.(2025)Phan, Gatti, Han, Li, Hu, Zhang, Zhang, Shaaban,
  Ling, Shi, Choi, Agrawal, Chopra, Khoja, Kim, Ren, Hausenloy, Zhang, Mazeika,
  Dodonov, Nguyen, Lee, Anderson, Doroshenko, Stokes, Mahmood, Pokutnyi, Iskra,
  Wang, Levin, Kazakov, Feng, Feng, Zhao, Yu, Gangal, Zou, Wang, Popov,
  Gerbicz, Galgon, and Schmitt}]{phan2025humanitysexam}
Long Phan, Alice Gatti, Ziwen Han, Nathaniel Li, Josephina Hu, Hugh Zhang, Chen
  Bo~Calvin Zhang, Mohamed Shaaban, John Ling, Sean Shi, Michael Choi, Anish
  Agrawal, Arnav Chopra, Adam Khoja, Ryan Kim, Richard Ren, Jason Hausenloy,
  Oliver Zhang, Mantas Mazeika, Dmitry Dodonov, Tung Nguyen, Jaeho Lee, Daron
  Anderson, Mikhail Doroshenko, Alun~Cennyth Stokes, Mobeen Mahmood, Oleksandr
  Pokutnyi, Oleg Iskra, Jessica~P. Wang, John-Clark Levin, Mstyslav Kazakov,
  Fiona Feng, Steven~Y. Feng, Haoran Zhao, Michael Yu, Varun Gangal, Chelsea
  Zou, Zihan Wang, Serguei Popov, Robert Gerbicz, Geoff Galgon, and Johannes
  Schmitt. 2025.
\newblock \href {http://arxiv.org/abs/2501.14249} {Humanity's last exam}.

\bibitem[{Schnitzler et~al.(2024)Schnitzler, Ho, Huang, Boudin, Sugawara, and
  Aizawa}]{morehopqamultihopreasoning}
Julian Schnitzler, Xanh Ho, Jiahao Huang, Florian Boudin, Saku Sugawara, and
  Akiko Aizawa. 2024.
\newblock \href {http://arxiv.org/abs/2406.13397} {Morehopqa: More than
  multi-hop reasoning}.

\bibitem[{Smith and Troynikov(2024)}]{Smith_Troynikov_2024}
Bradon Smith and Anton Troynikov. 2024.
\newblock \href {https://research.trychroma.com/evaluating-chunking}
  {\emph{Evaluating Chunking Strategies for Retrieval}}.

\bibitem[{Yang et~al.(2018)Yang, Qi, Zhang, Bengio, Cohen, Salakhutdinov, and
  Manning}]{hotpotqa}
Zhilin Yang, Peng Qi, Saizheng Zhang, Yoshua Bengio, William Cohen, Ruslan
  Salakhutdinov, and Christopher~D. Manning. 2018.
\newblock \href {https://doi.org/10.18653/v1/D18-1259} {{H}otpot{QA}: A dataset
  for diverse, explainable multi-hop question answering}.
\newblock In \emph{Proceedings of the 2018 Conference on Empirical Methods in
  Natural Language Processing}, pages 2369--2380, Brussels, Belgium.
  Association for Computational Linguistics.

\bibitem[{Zhuang et~al.(2024)Zhuang, Zhang, Cheng, Yang, Liu, Huang, Lin,
  Rajmohan, Zhang, and
  Zhang}]{Zhuang_Zhang_Cheng_Yang_Liu_Huang_Lin_Rajmohan_Zhang_Zhang_2024}
Ziyuan Zhuang, Zhiyang Zhang, Sitao Cheng, Fangkai Yang, Jia Liu, Shujian
  Huang, Qingwei Lin, Saravan Rajmohan, Dongmei Zhang, and Qi~Zhang. 2024.
\newblock \href {https://doi.org/10.18653/v1/2024.emnlp-main.199}
  {Efficientrag: Efficient retriever for multi-hop question answering}.
\newblock In \emph{Proceedings of the 2024 Conference on Empirical Methods in
  Natural Language Processing}, page 3392–3411, Miami, Florida, USA.
  Association for Computational Linguistics.

\end{thebibliography}


\begin{thebibliography}{0}
\expandafter\ifx\csname natexlab\endcsname\relax\def\natexlab#1{#1}\fi

\end{thebibliography}
